\documentclass[10pt]{article}

\usepackage{amsfonts,amsmath,amsthm}
\usepackage[numbers]{natbib}
\usepackage{algorithm,algorithmic}
\usepackage{bm,color}
\usepackage[dvipdfmx]{graphicx} 
\usepackage[margin=1in]{geometry}

\usepackage{macro_takeuchi}

\title{An Algorithmic Framework for Computing Validation Performance Bounds 
by Using Suboptimal Models
}

\date{February 10, 2014}

\author{
Yoshiki Suzuki \\
Department of Engineering \\
Nagoya Institute of Technology \\
Nagoya, Japan \\
\texttt{suzuki.mllab.nit@gmail.com} \\
\and 
Kohei Ogawa \\
Department of Engineering \\
Nagoya Institute of Technology \\
Nagoya, Japan \\
\texttt{ogawa.mllab.nit@gmail.com} \\
\and 
Yuki Shinmura \\
Department of Engineering \\
Nagoya Institute of Technology \\
Nagoya, Japan \\
\texttt{shinmura.mllab.nit@gmail.com} \\
\and 
Ichiro Takeuchi\thanks{Corresponding author} \\
Department of Engineering \\
Nagoya Institute of Technology \\
Nagoya, Japan \\
\texttt{takeuchi.ichiro@nitech.ac.jp} \\
}

\begin{document}

\maketitle

\clearpage
\begin{abstract}
Practical model building processes are often time-consuming because many different models must be trained and validated.
In this paper, we introduce a novel algorithm that can be used for computing the lower and the upper bounds of model validation errors without actually training the model itself.
A key idea behind our algorithm is using a side information available from a \emph{suboptimal model}.
If a reasonably good suboptimal model is available, our algorithm can compute lower and upper bounds of many useful quantities for making inferences on the unknown target model.
We demonstrate the advantage of our algorithm in the context of model selection for regularized learning problems.

 \vspace{.1in}
\vspace{.1in}
\end{abstract}

\clearpage

\section{Introduction}
\label{sec:introduction}
In practical model building processes, it is often required to train a large number of multiple different models.
Those models are usually evaluated based on a generalization performance measure such as the validation error (e.g., mis-classification error rate on a validation data set).
When the training algorithm of each of those models is formulated as an optimization problem, the entire model building process would be quite time-consuming.
It is, however, important to note that the final goal of model building is to find the single best model.
It means that we only need the validation error for the rest of the models and the model itself is not necessary.
If we could compute the validation error of a model without actually training it, model building processes would be much more efficient.

In this paper, we introduce a novel algorithm for a class of regularized learning problems.
Our algorithm can be used for computing the lower and the upper bounds of the validation error without actually solving the training optimization problem.
Instead of computing the validation error directly from the trained model itself, our algorithm uses a side information available from a \emph{suboptimal} model.
If we have a reasonably good suboptimal model that is sufficiently close to the target model, our algorithm can provide the bounds of the validation error.

Our algorithm is especially useful in model selection for regularized learning problems, where a sequence of models with various regularization parameters are trained and validated.
In this scenario, an already trained model with a certain regularization parameter can be used as the suboptimal model for our algorithm.
Then, we can compute the validation error bounds of other unknown models associated with other regularization parameters.
If the validation error lower bound of a model is larger than the smallest value obtained so far, we can skip training that model.

The basic idea behind our algorithm is computing a closed convex domain in the solution space in which we only know that the optimal solution exists, but the optimal solution itself is unknown.
If such a closed convex domain is available, it is often possible to compute the bounds of a quantity depending on the unknown optimal solution.
For a certain class of regularized learning problems, we show that such a domain can be easily derived and the bounds can be analytically computed based on a side information available from a suboptimal model.
This algorithmic trick is inspired from a recent study on \emph{safe screening} in the context of sparse modeling \cite{ElGhaoui12b}.

Our algorithm has a connection with recent studies on \emph{approximate regularization path}~\cite{Mairal12a,Giesen12a,Giesen12b}.
Its key property is the ability to compute the lower bounds of the objective values of the training optimization problems.
This property is useful for computing a regularization path with $\eps$-approximation guarantee.
In this context, our algorithm can be considered as a variant of such approximate regularization path algorithms.
Instead of bounding the objective values, our algorithm can compute an $\eps$-approximate regularization path in terms of validation errors, which is more useful for model selection purpose. 

Our main contribution in this paper is to implement the above idea in a general algorithmic framework, and show that it can be useful in many practical machine learning tasks.
Although we mainly focus on model selection for binary classification problems, our algorithm can be applied to any learning problems defined with a convex loss function and an $L_2$ regularizer.
It can compute the lower and the upper bounds of many useful quantities for making inferences on unknown target models. 
To the best of our knowledge, there are no other previously known algorithms that can compute practically useful bounds for various types of model evaluation performances.

 \section{Problem Setup and Basic Idea}
\label{sec:2}

\paragraph{Notations}
For any natural number $n$, we define $[n]:= \{1, \ldots, n\}$.
A real $n$-vector is denoted as $v \in \RR^n$ and $v^\top$ indicates the transpose of the vector.
Unless otherwise stated, $\|\cdot\|$ is a Euclidean norm. 

\paragraph{Problem setup}
Let us denote the training set as $\{(x_i, y_i)\}_{i \in [n]}$, where $x_i \in \cX$ is the input vector in the input space $\cX$ and $y_i \in \{\pm 1\}$ is the binary class label.
Let $\phi: \cX \to \cF$ be a feature map associated with a kernel $K$.
We consider a linear model in the feature space $\cF$ in the following form:
\begin{align*}
 f(x) = \phi(x)^\top w, 
\end{align*}
where $w \in \cF$ is the vector of coefficients.
For simplicity, we denote $\phi_i := \phi(x_i), i \in [n]$.
We consider the following class of $L_2$ regularized convex learning problems: 
\begin{align}
 \label{eq:original.problem}
 w^*_{C} := 
 \arg \min_{w \in \cF}
 ~
 \frac{1}{2}
 \| w\|^2
 +
 C
 \sum_{i \in [n]}
 \ell(y_i, \phi_i^\top w),
\end{align}
where $\frac{1}{2} \| w\|^2$ is an $L_2$ regularization term, $\ell$ is a convex loss function, and $C > 0$ is the regularization parameter for controlling the balance between the two terms.
We denote the optimal solution as $w^*_C$ in order to clarify that it is the optimal solution associated with the regularization parameter $C$.
With a slight abuse of notation, we use the following simplified notation when there is no ambiguity:
\begin{align*}
 \ell_i(w) := \ell(y_i, \phi_i^\top w).
\end{align*}

\paragraph{Basic idea}
In this paper, we develop a general algorithmic framework for computing the lower and the upper bounds of the inner product $\theta^\top w^*_C$ for an arbitrary vector $\theta \in \cF$ \emph{without actually solving the optimization problem} for $w^*_C$.
We denote the lower and the upper bounds as $b_{lo}(\theta^\top w^*_{\rm C})$ and $b_{up}(\theta^\top w^*_{\rm C})$, respectively, i.e.,
\begin{align*}
 b_{lo}(\theta^\top w^*_{\rm C})
 \le
 \theta^\top w^*_C
 \le
 b_{up}(\theta^\top w^*_{\rm C}).
\end{align*}
We will demonstrate that this framework is quite useful in many practical machine learning tasks.

If we have a validation data set $\{(x^\prime_i, y^\prime_i)\}_{i \in [n^\prime]}$ for a binary classification problem with $x^\prime_i \in \cX$ and $y^\prime_i \in \{\pm 1\}$, the mis-classification error rate
\begin{align*}
 \frac{1}{n^\prime}
 \sum_{i \in [n^\prime]}
 I \left\{ y^\prime_i \neq {\rm sgn}( \phi_i^{\prime \top} w^*_{C}) \right\}
 \end{align*}
can be bounded from below and above by
\begin{align}
 \label{eq:misclassification.rate.LB}
 \frac{1}{n^\prime} \biggl(
 \sum_{i: y_i^\prime = +1}
 I \{b_{up}(\phi_i^{\prime \top} w^*_{C}) < 0 \}
 + 
 \sum_{i: y_i^\prime = -1}
 I \{b_{lo}(\phi_i^{\prime \top} w^*_{C}) > 0 \}
 \biggr),
\end{align}
and
\begin{align}
 \label{eq:misclassification.rate.UB}
 1 - \frac{1}{n^\prime}
 \biggl(
 \sum_{i: y_i^\prime = +1}
 I\{b_{lo}(\phi_i^{\prime \top} w^*_C) > 0\}
 + \sum_{i: y_i^\prime = -1}
 I\{b_{up}(\phi_i^{\prime \top} w^*_C) < 0\}
 \biggr),
\end{align}
respectively, where $I(\cdot)$ is the indicator function, ${\rm sgn}(\cdot)$ is the sign function, and $\phi_i^\prime := \phi(x_i^\prime)$.

Although we focus in this paper on the problem of computing validation error bounds for binary classification problems, our framework for bounding $\theta^\top w^*_C$ is far more general.
It can be used for computing the lower and the upper bounds of many useful quantities for validation, inference and prediction on various models.

Our basic algorithmic idea for computing the bounds of $\theta^\top w^*_C$ is as follows. 
Suppose that we only know that the optimal solution $w^*_C$ is somewhere in a closed convex domain $\cS \in \cF$, but we do not know the optimal solution $w^*_C$ itself.
In such a case, the lower and the upper bounds of $\theta^\top w^*_C$ can be obtained by solving the following minimization and maximization problems:
\begin{subequations} 
\label{eq:basic.idea}
\begin{align}
 b_{lo}(\theta^\top w^*_C)
 &:= \min_{w \in \cS}
 ~
 \theta^\top w,
 \\
 b_{up}(\theta^\top w^*_C)
 &:= \max_{w \in \cS}
 ~
 \theta^\top w. 
\end{align}
\end{subequations}
We later show that, for the class of regularized learning problems in \eq{eq:original.problem}, we can easily find such a closed convex domain $\cS$, and the lower and the upper bounds in the forms of \eq{eq:basic.idea} can be analytically computed.

This algorithmic trick is inspired from a recent study on \emph{safe screening} in the context of sparse modeling \cite{ElGhaoui12b}.
Safe screening enables to identify and screen out a part of the sparse model coefficients which turn out to be 0 at the optimal solution \emph{before} actually training the model.
Although our problem setup and goal are totally different, some of the algorithmic and proof techniques developed in \cite{ElGhaoui12b} and the subsequent studies \cite{Xiang12a,Xiang12b,Dai12a,Wang12a,Ogawa13a,Wang13c,Wang13d,Wu13a,Wang13a,Wang13b,Ogawa14a} are useful for our algorithm development (see Appendix~\ref{theo:App.to.LASSO} for more discussion on the relation between our approach and safe screening).

\section{Bounds by Suboptimal Models}
\label{sec:3}
In this section we present our main results.
In Theorem~\ref{theo:main}, we first describe our general result for computing the lower and the upper bounds of a quantity depending on the unknown optimal solution. 
In Theorem~\ref{theo:bounds.model.selection}, we focus on model selection scenario for regularized learning problems, where we derive the lower and the upper bounds represented as the functions of the regularization parameter $C$.

\begin{theo}
 \label{theo:main}
 Let $\tilde{w} \in \cF$ be an arbitrary vector in the feature space. 
 Then,
 \begin{align}
  \label{eq:ball}
  w^*_C \in \cS := \{ w ~|~ \| w - m \| \le r\}, 
 \end{align}
 i.e., the optimal solution $w_C^*$ is in the ball $\cS$ whose center $m \in \cF$ and the radius $r > 0$ are defined as
\begin{subequations}
  \label{eq:m.and.r}
 \begin{align}
  \label{eq:m}
  m &:= \frac{1}{2} \Bigl(\tilde{w} - C \sum_{i \in [n]} \nabla \ell_i(\tilde{w}) \Bigr), 
  \\
  \label{eq:r}
  r &:= \frac{1}{2} \Bigl\|\tilde{w} + C \sum_{i \in [n]} \nabla \ell_i(\tilde{w})\Bigr\|,
 \end{align}
 \end{subequations}
 where $\nabla \ell_i(\tilde{w}) \in \cF$ is the gradient vector of $\ell_i$ at $w = \tilde{w}$ when $\ell_i$ is differentiable at $\tilde{w}$, while it is an arbitrary subgradient vector of $\ell_i$ at $w = \tilde{w}$ when $\ell_i$ is non-differentiable at $\tilde{w}$.

 It indicates that, for any $\theta \in \cF$, the inner product $\theta^\top w^*_C$ are bounded as 
\begin{align*}
 \theta^\top m  - \| \theta \| r
 \le
 \theta^\top w^*_C
 \le 
 \theta^\top m + \| \theta \| r,
\end{align*}
 i.e., the lower and the upper bounds are written as
 \begin{subequations}
  \label{eq:main.theo.bounds}
   \begin{align}
  b_{lo}(\theta^\top w^*_C)
  &:=
  \theta^\top m  - \| \theta \| r, \\
  b_{up}(\theta^\top w^*_C)
  &:=
  \theta^\top m  + \| \theta \| r.
   \end{align}
 \end{subequations}
 \end{theo}
 The proof of Theorem~\ref{theo:main} is presented in Appendix~\ref{app:proofs}.
 
 Theorem~\ref{theo:main} is quite general because an arbitrary \emph{suboptimal solution} $\tilde{w} \in \cF$ can be used for computing the bounds.
 However, it is important to note that, if we do not have a reasonably \emph{good} suboptimal solution, the bounds in \eq{eq:main.theo.bounds} could be quite loose and practically useless.
 We could roughly say that the bounds are tight when the suboptimal solution $\tilde{w}$ is close to the optimal solution $w^*_C$ (see \S~\ref{sec:5} for simple simulation results on this issue).
 The tightness of the bounds also depends on the curvature of the objective function\footnote{
 Since Theorem~\ref{theo:main} tells that the solution is in a ball, the tightness of the bounds are closely related to the radius $r$. 
 When the loss function $\ell_i$ is differentiable and the optimal solution $w^*_C$ itself is used as the suboptimal model $\tilde{w}$ in Theorem~\ref{theo:main}, we could see that the radius is 0, i.e.,
 $r = \frac{1}{2} \| w^*_C + C \sum_{i \in [n]} \nabla \ell_i(w^*_C)\| = 0$,
 and  the bounds in \eq{eq:main.theo.bounds} are exact.}.

 The following special case is very useful in the context of model selection for regularized learning problems.
 If we regard the optimal solution with a different regularization parameter $\tilde{C} > 0$ as the suboptimal solution in Theorem~\ref{theo:main}, i.e., if we set $\tilde{w} := w^*_{\tilde{C}}$ for a certain $\tilde{C} > 0$, the lower and the upper bounds of $\theta^\top w^*_C$ are represented in simple interpretable forms.
 \begin{theo}
  \label{theo:bounds.model.selection}
  Let
  $w^*_{\tilde{C}}$
  be the optimal solution of the problem
  \eq{eq:original.problem}
  for a regularization parameter
  $\tilde{C} > 0$.
  Then,
  for any
  $\theta \in \cF$,
  the lower and the upper bounds of the inner product 
  $\theta^\top w^*_C$ 
  are written as
  \begin{subequations}
   \label{eq:bounds.model.selection}
  \begin{align}
  \label{eq:low.bounds.model.selection}
  b_{lo}(\theta^\top w^*_C) 
  &
  =
  \mycase{
   \frac{1}{2}
   (\theta^\top w^*_{\tilde{C}} + \| \theta \| \| w^*_{\tilde{C}} \|) 
   + 
   \frac{C}{2 \tilde{C}}
   (\theta^\top w^*_{\tilde{C}} - \| \theta \| \| w^*_{\tilde{C}} \|)
   & \text{if~}\tilde{C} < C,
   \\
   \frac{1}{2}
   (\theta^\top w^*_{\tilde{C}} - \| \theta \| \| w^*_{\tilde{C}} \|)
   + 
   \frac{C}{2 \tilde{C}}
   (\theta^\top w^*_{\tilde{C}} + \| \theta \| \| w^*_{\tilde{C}} \|)
   & \text{if~}\tilde{C} > C,
   }
  \\
  \label{eq:upp.bounds.model.selection}
  b_{up}(\theta^\top w^*_C) 
  &
  =
  \mycase{
   \frac{1}{2}
   (\theta^\top w^*_{\tilde{C}} - \| \theta \| \| w^*_{\tilde{C}} \|) 
   +
   \frac{C}{2 \tilde{C}}
   (\theta^\top w^*_{\tilde{C}} + \| \theta \| \| w^*_{\tilde{C}} \|)
   & \text{if~}\tilde{C} < C,
   \\
   \frac{1}{2}
   (\theta^\top w^*_{\tilde{C}} + \| \theta \| \| w^*_{\tilde{C}} \|)
   + 
   \frac{C}{2 \tilde{C}}
   (\theta^\top w^*_{\tilde{C}} - \| \theta \| \| w^*_{\tilde{C}} \|)
   & \text{if~}\tilde{C} > C.
}
  \end{align}
  \end{subequations}
 \end{theo}
 The proof of
 Theorem~\ref{theo:bounds.model.selection}
 is presented in Appendix~\ref{app:proofs}.

 Interestingly,
 the bounds in
 \eq{eq:bounds.model.selection}
 are represented as the functions of the regularization parameter $C$.
 It implies that,
 once we compute the optimal solution associated with a regularization parameter $\tilde{C}$, 
 we can obtain a continuum path of the lower and the upper bounds of
 $\theta^\top w^*_C$ 
 parametrized by the regularization parameter $C$.
 The following corollary describes a few important properties of these parametrized bounds.
 \begin{coro}
  (i) The lower bound
  \eq{eq:low.bounds.model.selection}
  is monotonically decreasing with $C$
  for
  $C > \tilde{C}$,
  and
  monotonically increasing with $C$
  for
  $C < \tilde{C}$.
  Similarly,
  the upper bound
  \eq{eq:upp.bounds.model.selection}
  is monotonically increasing with $C$
  for
  $C > \tilde{C}$,
  and
  monotonically decreasing with $C$
  for
  $C < \tilde{C}$.
  (ii)
  Furthermore,
  the lower and the upper bounds converge to 
  $\theta^\top w^*_{\tilde{C}}$
  as $C$ approaches to $\tilde{C}$.
 \end{coro}
 \begin{proof}
  The part (i) can be easily proved by noting that
  \begin{align*}
   \theta^\top w^*_{\tilde{C}} - \|\theta\| \|w^*_{\tilde{C}}\| \le 0
   \text{~and~}
   \theta^\top w^*_{\tilde{C}} + \|\theta\| \|w^*_{\tilde{C}}\| \ge 0
  \end{align*}
  from the Cauchy-Schwartz inequality.
  For the part (ii), it is also clear to note that
  \begin{align*}
   \lim_{C \to \tilde{C}} ~ b_{lo}(\theta^\top w^*_{\tilde{C}}) 
   = \lim_{C \to \tilde{C}} ~ b_{up}(\theta^\top w^*_{\tilde{C}})
   = \theta^\top w^*_{\tilde{C}}.
  \end{align*}
 \end{proof}

 \paragraph{Bounds in the Intersection of Two Balls}
 If we have two suboptimal models 
 $\tilde{w}_1, \tilde{w}_2 \in \cF$,
 the optimal solution
 $w^*_C$
 is in the intersection of the two corresponding balls 
 $\cS_1$
 and 
 $\cS_2$. 
 Since the intersection is smaller than each ball by definition,
 the bounds
 $\min_{w \in \cS_1 \cap \cS_2} \theta^\top w$
 and
 $\max_{w \in \cS_1 \cap \cS_2} \theta^\top w$
 are tighter than those obtained from a single ball.
 If the two balls are denoted as
 $\cS_1 := \{w ~|~ \|w - m_1\| \le r_1\}$
 and 
 $\cS_2 := \{w ~|~ \|w - m_2\| \le r_2\}$,
 using the Lagrange multiplier methods (and tedious algebraic computation), 
 the lower and the upper bounds in the intersection are computed as follows:
  \begin{align*}
   \min_{w \in \cS_1 \cap \cS_2} \theta^\top w = \mycase{
   \displaystyle\min_{w \in \cS_1} \theta^\top w, & 
   \text{~if~} 
   \frac{- \theta^\top \alpha}{\|\theta\|\|\alpha\|} < \frac{\beta - \|\alpha\|}{r_1}, \\
   \displaystyle\min_{w \in \cS_2} \theta^\top w, & 
   \text{~if~} 
   \frac{\beta}{r_2} < \frac{-\theta \alpha}{\|\theta\|\|\alpha\|}, \\
   \theta^\top \gamma - \delta \left(\|\theta\|^2 - \frac{\|\theta^\top \alpha\|^2}{\|\alpha\|^2} \right)^{\frac{1}{2}}, 
   &
   \text{otherwise}. 
   }
   \\
   \max_{w \in \cS_1 \cap \cS_2} \theta^\top w = \mycase{
   \displaystyle\max_{w \in \cS_1} \theta^\top w, & 
   \text{~if~} 
   \frac{\theta^\top \alpha}{\|\theta\|\|\alpha\|} < \frac{\beta - \|\alpha\|}{r_1}, \\
   \displaystyle\max_{w \in \cS_2} \theta^\top w, & 
   \text{~if~} 
   \frac{\beta}{r_2} < \frac{\theta \alpha}{\|\theta\|\|\alpha\|}, \\
   \theta^\top \gamma + \delta \left(\|\theta\|^2 - \frac{\|\theta^\top \alpha\|^2}{\|\alpha\|^2} \right)^{\frac{1}{2}}, 
   &
   \text{otherwise},
   }
  \end{align*}
  where
  $\alpha := m_1 - m_2$,
  $\beta := (\|\alpha\|^2 + r_2^2 - r_1^2)/(2 \alpha)$,
  $\gamma := m_2 + \beta \alpha/\|\alpha\|$,
  $\delta := (r_2^2 - \beta^2)^{1/2}$.
  The same technique has been also used in the context of safe screening~\cite{Ogawa14a}.
  Although it is possible to consider the intersection of more than two balls,
  it requires much more tedious algebraic computations.

  In a part of the experiments (see \S~\ref{sec:5}),
  we use a simple but useful trick using the above intersection.
  When we have a suboptimal solution
  $\tilde{w} \in \cF$,
  we can make use of the center $m \in \cF$ in
  \eq{eq:m}
  as another suboptimal solution,
  and consider the intersection of the resulting two balls.
  We show in Lemma~\ref{lemm:Recursive.IT} in Appendix
  that the area of the intersection is less than the half of the original ball,
  meaning that the new bounds could be much tighter than the original ones.
  
\paragraph{Kernelization}
The bounds \eq{eq:bounds.model.selection} in Theorem~\ref{theo:bounds.model.selection} can be \emph{kernelized}, i.e., what we need to compute in (possibly infinite-dimensional) feature space $\cF$ is only inner products which can be computed by using the associated kernel function $K$.
The bounds \eq{eq:main.theo.bounds} in Theorem~\ref{theo:main} can be also kernelized if $\nabla \ell_i(w), i \in [n],$ can be kernelized.
\section{Applications}
\label{sec:4}
In this section, we present several practical machine learning tasks in which our algorithmic framework for computing bounds is useful.

\subsection{Efficient Model Selection}
\label{subsec:app.efficient.model.selection}
Let us first discuss how our bound computation framework can be used 
in ordinary model selection scenario.
We consider model selection problems for an $L_2$ regularized convex
learning problem in the form of \eq{eq:original.problem}.
We consider a common situation that two separate training and validation
sets are available for training and model selection, respectively.
Here, our task is to select the best regularization parameter $C$
that yields the smallest mis-classification error rate on the validation set 
among a given list of the candidates 
$C_1, \ldots, C_T$. 
In general,
we need to solve all the $T$ optimization problems
for finding the best one\footnote{
For a certain class of problems, one can compute the exact path of the
optimal solutions for the entire range of $C$, which is referred to as
\emph{regularization path}~\cite{HasRosTibZhu04}.
Regularization path computation is possible only for a limited class of problems
(e.g., it can be computed for an SVM, but not for logistic regression).
In addition, regularization path computation is known to be numerically unstable, and does not scale well.}~\footnote{
It is also beneficial in practice to use \emph{warm-start} approaches
\cite{DecWag00} when solving a sequence of optimization problems.
For simplicity, 
we do not take into account the possible advantage of
warm-start approach in our discussion here.}.

We can use the bounds in Theorem~\ref{theo:bounds.model.selection}
for making the model selection problem more efficient.
If we have already computed a solution
for a certain $C_{\tilde{t}}, \tilde{t} \in [T]$, 
we can use this solution as the suboptimal solution
$\tilde{w}$.
Then, 
the lower and the upper bounds of the mis-classification error rate
in
\eq{eq:misclassification.rate.LB}
and
\eq{eq:misclassification.rate.UB},
respectively, 
are computed for the remaining candidates.
We can make use of the lower bounds for skipping some of the $T$ training tasks, i.e., if the lower bound of the validation error for a certain $C_t$ is larger than the smallest validation error obtained so far, we can skip training that model. 
In addition, these lower and upper bounds are helpful to decide which model should be trained in the next step\footnote{
In our experiments in \S~\ref{sec:5}, we just selected the $C_t$ whose validation error lower bound is smallest.
There are, however, many other possible approaches.
For example, we can select the $C_t$ whose uncertainty (the difference between the upper and the lower bounds) are largest. 
See Bayesian optimization for hyperparameter search~\cite{Sneok12a} for detailed discussion on this issue.
}.
A summary of the efficient model selection procedure is described in Algorithm~\ref{alg:EMS}.

\renewcommand{\algorithmicrequire}{\textbf{Input:}}
\renewcommand{\algorithmicensure}{\textbf{Output:}}
\begin{algorithm}[H]
\caption{Efficient Model Selection Algorithm} 
\label{alg:EMS}
 \begin{algorithmic}[1]
  \REQUIRE training set $\cD_{tr}$, validation set $D_{va}$, a list of regularization parameters $\{C_t\}_{t \in [T]}$
  \ENSURE the optimal solution $w^*_{C_{\rm best}}$
  
  \STATE $\veps^\ell_t \lA 0.0, \veps^u_t \lA 1.0 ~ \forall t \in [T]$
  \STATE $\veps_{\rm best} \lA 1.0$, $t_{\rm best} \lA 1$, $\cT \lA [T] \setminus \{1\}$
  \WHILE {$\exists~t \in \cT ~ \text{such that}~ \veps_t^\ell < \veps_{\rm best} $}
  \STATE ${\hat{t}} \lA$ \textsf{ChooseNextC}($\{\veps_t^\ell, \veps_t^u\}_{t \in \cT}$)
  \STATE $w^*_{C_{\hat{t}}} \lA \textsf{TrainModel}(\cD_{tr}, C_{\hat{t}}) $
  \STATE $\veps_{\hat{t}} \lA \textsf{ComputeValidError}(w^*_{C_{\hat{t}}}, \cD_{va})$
  \STATE $\cT \lA \cT \setminus \{\hat{t}\}$
  \IF { $\veps_{{\hat{t}}} < \veps_{\rm best} $ }
  \STATE $\veps_{\rm best} \lA \veps_{\hat{t}}$, $w^*_{C_{\rm best}} \lA w^*_{C_{\hat{t}}}$
  \ENDIF
  \FOR { $t \in \cT$ }
  \STATE $\{\tilde{\veps}_t^\ell, \tilde{\veps}_t^u\} \lA \textsf{ComputeValidErrorBounds}(w^*_{C_{\hat{t}}}, \cD_{va}, C_t)$
  \STATE $\veps_t^\ell \lA \max\{\veps_t^\ell, \tilde{\veps}_t^\ell\}$, $\veps_t^u \lA \min\{\veps_t^u, \tilde{\veps}_t^u\}$, 
  \ENDFOR
  \ENDWHILE
 \end{algorithmic}
\end{algorithm}
In Algorithm~\ref{alg:EMS}, 
\textsf{ChooseNextC}
is a function for selecting one of the remaining regularization parameter
$C_t \in \cT$
for the next step.
The basic idea here is to select the candidate with which the validation error is expected to be smallest. 
In this paper,
we simply select
$\arg \min_{t \in \cC} \veps^\ell_t$
as the next candidate. 
The function 
\textsf{TrainModel}
is used for training the model with the specified regularization parameter.
Any specific solvers or general convex optimization tools can be used for this function.
The function
\textsf{ComputeValidError}
computes the validation error
based on a given solution.
The function 
\textsf{ComputeValidErrorBounds}
computes the validation error bounds at the specified regularization parameter
based on a given solution.

\subsection{Exact and approximate model selection}
\label{subsec:app.exact.approximate.model.selection}

\paragraph{Exact model selection}
Although it is common to select the regularization parameter among the finite list of the candidates as we discussed in
\S~\ref{subsec:app.efficient.model.selection},
it would be better if we could 
find the best possible regularization parameter
that exactly minimizes the validation error in the continuous range of 
$C \in [C_{\rm min}, C_{\rm max}]$\footnote{
Ideally, we should select the best $C$ from $(0, \infty)$.
But it is practically difficult except some special cases.
We thus consider selecting $C$ from an interval between 
$C_{\rm min}$ and $C_{\rm max}$.
}.
For the class of $L_2$-regularized convex learning problems in the form
of \eq{eq:original.problem}, such \emph{exact model selections} are possible
because we can compute the lower bounds of the validation errors for the 
continuum of the regularization parameters
$C \in [C_{\rm min}, C_{\rm max}]$.

Suppose that we have already solved an optimization problem
\eq{eq:original.problem}
for a certain $\tilde{C} < C$,
and denote the solution as
$w^*_{\tilde{C}}$.
Then,
for an input 
$x_i^\prime$
in the validation set,
the following rules can be obtained from \eq{eq:bounds.model.selection}:
\begin{align*}
 &
 \tilde{C} < C <
 \frac{
 \|\phi_i^{\prime}\|  \|w^*_{\tilde{C}}\| + \phi_i^{\prime \top} w^*_{\tilde{C}}
 }{
 \|\phi_i^{\prime}\|  \|w^*_{\tilde{C}}\| - \phi_i^{\prime \top} w^*_{\tilde{C}}
 } \tilde{C}
 ~~\Rightarrow~~
 \phi_i^{\prime \top} w^*_C > 0.
 \\
 &
 \tilde{C} < C < \frac{
 \|\phi_i^{\prime}\|  \|w^*_{\tilde{C}}\| - \phi_i^{\prime \top} w^*_{\tilde{C}}
 }{
 \|\phi_i^{\prime}\|  \|w^*_{\tilde{C}}\| + \phi_i^{\prime \top} w^*_{\tilde{C}}
 }
 \tilde{C}
 ~~\Rightarrow~~
 \phi_i^{\prime \top} w^*_C < 0.
\end{align*}
Using the above rules, 
we can compute the lower bounds of the validation errors
\eq{eq:misclassification.rate.LB}
as a function of
$C \in [C_{\rm min}, C_{\rm max}]$.
It means that we can exactly identify a sequence of the regularization parameter values
at which the validation error changes by $1/n^\prime$.
In other words, 
we can trace all the change points of the validation error along
$C \in [C_{\rm min}, C_{\rm max}]$.
%

\paragraph{Model selection with approximation guarantee}
The above exact model selection can be relaxed
so that it allows to have an $\veps$-approximation error,
i.e.,
we can compute a sequence of the models among which
there exists a solution whose validation error is within $\eps$ from 
the minimum possible value in 
$C \in [C_{\rm min}, C_{\rm max}]$.
For example,
if we set
$\eps$
such that
$\lfloor n^\prime \eps \rfloor = 5$, 
then
we can compute the sequence of points in $[C_{\rm min}, C_{\rm max}]$
at which the validation error changes by $5/n^\prime$.

This model selection scheme
can be considered as a valiant of
\emph{approximate regularization path}~\cite{Mairal12a,Giesen12a,Giesen12b}.
The key property of these approximate regularization path algorithms is
computing the path of solutions
with which the approximation error of the objective function values
are bounded by $\eps$.
In our approach,
we can control the approximation error of validation performances, 
which is more useful for model selection purpose. 

\subsection{Fast leave-one-out cross validation}
Next,
we propose to use our bounds
for efficient computation of leave-one-out cross-validation (LOOCV)
in binary classification problems.
With a slight abuse of notation,
let us denote the optimal solution trained with all the instances as
$w^*_{\rm all}$,
while 
the optimal solution obtained after picking out an instance $(x_j, y_j)$ as
\begin{align}
 \label{eq:LOOCV.computation}
 w^*_{(-j)} := \arg \min_{w \in \cF}
 ~
 \frac{1}{2} \|w\|^2 + C \sum_{i \neq j} \ell(y_i, \phi_i^\top w).
\end{align}
Then, 
the LOOCV error is written as
\begin{align*}
 \frac{1}{n} \bigl(
 \sum_{i:y_j = +1} I(\phi_j^\top w^*_{(-j)} < 0)
 +
 \sum_{i:y_j = -1} I(\phi_j^\top w^*_{(-j)} > 0)
\bigr).
\end{align*}
Our idea here is to compute the bounds of
$\phi_j^\top w^*_{(-j)}$
using $w^*_{\rm all}$ 
as the suboptimal solution for our algorithm.
An advantage of this simple approach is that,
once we compute
$w^*_{\rm all}$,
it can be used as the suboptimal solution for bounding all the $n$ inner products 
$\phi_j^\top w^*_{(-j)} ~ \forall j \in [n]$.
If
$\phi_j^\top w^*_{(-j)}$
could be bounded from above or below 0, 
we do not have to compute the optimal $w^*_{(-j)}$.
If there are many such instances, 
the LOOCV computation process would be quite efficient. 

\subsection{Logistic model inference by SVM}
\label{subsec:LR.inf.SVM}
Our final application is to make inferences on a model based on a
suboptimal model trained by a different learning algorithm.
Specifically, we make inferences on a logistic regression model by using
the SVM solution trained with the same data set.
Logistic regression is especially important and popularly used in
biomedical research because the model output and model
coefficients are interpreted as the log odds and log odds ratios,
respectively.
On the other hand, SVM is more popularly used in large-scale machine
learning  and pattern recognition problems partly because it tends to
produce better classification performances and there are many efficient
algorithms and solvers that are applicable to large-scale data sets.
It is thus important to know how SVM solutions can be useful for
inferences on logistic regression models.
%

Our goal is to make inferences on the solution of the following $L_2$ regularized logistic regression model
\begin{align*}
 w_{\rm LR}^* := \frac{1}{2} \|w\|^2 + C \sum_{i \in [n]} \log \left(1 + \exp(- y_i x_i^\top w)\right)
\end{align*}
by using the suboptimal solution
$\tilde{w} := w^*_{\rm svm}$
given by
\begin{align*}
 w_{\rm svm}^* := \frac{1}{2} \|w\|^2 + C \sum_{i \in [n]} \max\{0, 1 - y_i x_i^\top w\}.
\end{align*}
Here, we only consider a linear model, i.e., the feature space $\cF$ is
$d$-dimensional Euclidean space $\RR^d$.

Our first interest is in each coefficient of the logistic model solution
$(w^*_{\rm LR})_ j$
for
$j \in [d]$
because it represents the log odds ratio of the $j^{\rm th}$ feature.
Using Theorem~\ref{theo:main},
we can compute the lower and the upper bounds of 
$(w^*_{\rm LR})_ j$
by bounding the inner product 
$e(j)^\top w^*_{{\rm LR}}$,
where $e(j)$ is the $j^{\rm th}$ coordinate unit vector.

Given the input of a new instance $x_{\rm new}$ (e.g., when a new patient profile is provided), 
our second task is to make an inference on the log odds of the instance.
We can compute the lower and the upper bounds of the log odds
by bounding the inner product
$x_{\rm new}^\top w^*_{\rm LR}$
using Theorem~\ref{theo:main}.

 \section{Numerical Experiments}
\label{sec:5}
In this section, we illustrate the effectiveness of our approach by
numerical experiments. 
We used 12 benchmark data sets listed in
Table~\ref{tab:dataset}.
%
We used SVM and Logistic Regression (LR)
as the two examples of regularized learning problems
in the form of \eq{eq:original.problem}.
LIBSVM~\cite{Chang11a}
and 
LIBLINEAR~\cite{Fan08b}
were used as the SVM and LR solvers\footnote{
The former provides kernel SVM solver, while the latter provides a linear SVM and a linear LR solvers.
}.

We report the results on both linear and nonlinear cases\footnote{
For nonlinear LR,
we just used basis expansion approach with Gaussian RBF, and the
optimization is conducted by linear LR solver in LIBLINEAR.}.
In nonlinear cases,
Gaussian kernel
$K(x_i, x_j) = \exp(- \gamma \|x_i - x_j\|^2)$
with
$\gamma = 1/d$
is used.
%
%

\begin{table}[t]
 \centering
 \caption{Datasets used in \S~\ref{sec:5}.}
\label{tab:dataset}
{\small 
 \begin{tabular}{rl|r|r} 
\multicolumn{1}{c}{ID} & Dataset & $n~~~$ & $d~$ \\
\hline
\hline
BCP :& BreastCancerPrognostic & 194  & 33 \\
\hline
PKS :& Parkinsons & 195 & 22 \\
\hline
SPH :& SPECTHeart & 267 & 44 \\ 
\hline
LVD :& Liver-Disorders & 345 & 6 \\
\hline
ION :& Ionosphere & 351 & 33 \\
\hline
BCI :& BrainComputerInterface & 400 & 117 \\
\hline
BCD :& BreastCancerDiagnostic & 569 & 30 \\
\hline
AUS :& Australian & 690 & 14 \\
\hline
G2C :& g241c & 1,500 & 241 \\
\hline
G2N :& g241n & 1,500 & 241 \\
\hline
SPM :& Spambase & 4,601 & 57 \\
\hline
MGT :& MAGICGammaTelescope & 19,020 & 10 \\
 \end{tabular}
}
\end{table}

\begin{figure}[htbp]
 \begin{center}
  \begin{tabular}{cc}
   \includegraphics[width=0.45\textwidth]{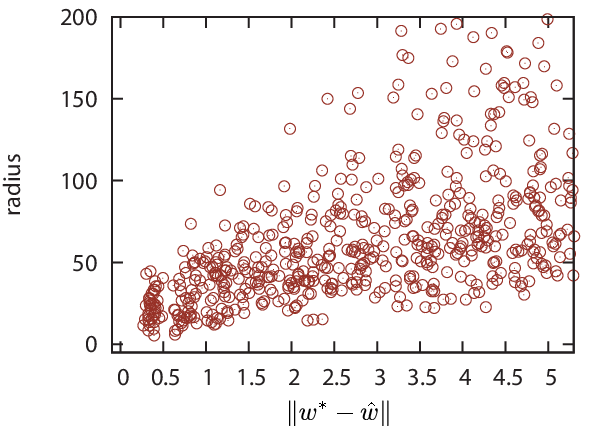} &
   \includegraphics[width=0.45\textwidth]{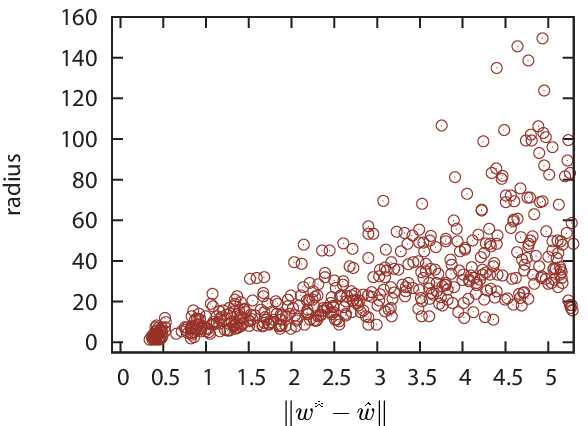} \\
   {\footnotesize (a) BCD data} &
   {\footnotesize (b) PKS data} 
  \end{tabular}
   \caption{The simulation results for understanding the effects of the suboptimal model on two data sets.}
  \label{fig:goodness.suboptimal}
 \end{center}
\end{figure}

\paragraph{Goodness of suboptimal solutions}
We conducted simple numerical simulations for understanding the effect of
the choice of suboptimal solutions. 
\figurename~\ref{fig:goodness.suboptimal}
shows the simulation results of linear LR on two data sets.
Here,
we randomly generated 1000 suboptimal solutions
by adding a Gaussian noise to the optimal solution. 
The x-axis denotes the distance from the optimal solution
$\| \tilde{w} - w^* \|$, 
while 
the y-axis denotes the tightness of the bounds in
\eq{eq:main.theo.bounds}
measured by the radius
$r$
in
\eq{eq:r}.
The results indicate that tighter bounds can be obtained 
as the selected suboptimal solutions approach to the optimal solution. 

\paragraph{Efficient model selection}
We examined the efficiency of the model selection strategy discussed in
\S~\ref{subsec:app.efficient.model.selection}.
Our task  is to find the best regularization parameter
$C$
among
$T = 501$
candidates
$\{C_1, \ldots, C_T\}$
evenly allocated
between 0.01 and 10000
in logarithmic scale.
The basic strategy is to sequentially training the models based on the
validation error bounds obtained so far. 
At each step, we just selected the model that has the smallest
validation error lower bound in \eq{eq:misclassification.rate.LB}.
In this experimental setup, 
we have multiple trained models that can be used as the suboptimal models.
We thus used the closest two models (one with smaller $C$ and the other with larger $C$)
as the suboptimal models,
and employed the intersection approach discussed in
\S~\ref{sec:3}.
\figurename~\ref{fig:exp.ems}
shows the validation error bounds after the last step where we could find
the best regularization parameter.
Table~\ref{tbl:EMS1}
shows how many training optimization problems were solved before finding
the best one.
The results indicate that
the best regularization parameter can be found 
without solving all the $T = 501$ optimization problems.
%

\begin{table}[h]
 \caption{The number of optimization problems solved before finding the
 best regularization parameter (among the 501 models).}
\label{tbl:EMS1}
 \centering
 \begin{tabular}{c}
  {\small
  \begin{minipage}{0.475\hsize}
  \vspace*{2mm}
   \centering
\text{Linear Model}\\
\vspace*{2mm}
   \begin{tabular}{c|rr}
	\hline 
	Data & LR~~~ & SVM~~ \\
	\hline 
	BCP & 421/501 & 196/501 \\ 
	LVD & 274/501 & 122/501 \\ 
	ION &  98/501 & 151/501 \\
	G2C & 337/501 & 99/501 \\
	\hline
   \end{tabular}
  \end{minipage}
  
  \begin{minipage}{0.475\hsize}
  \vspace*{2mm}
   \centering
   \text{Nonlinear Model}\\
\vspace*{2mm}
   \begin{tabular}{c|rr}
	\hline 
	Data & LR~~~ & SVM~~ \\
	\hline
	BCP & 321/501  & 56/501 \\ 
	PKS & 366/501 & 58/501 \\ 
	SPH & 381/501 & 74/501 \\
	BCD & 336/501 & 54/501  \\
	\hline
   \end{tabular}
  \end{minipage}
}
 \end{tabular}
\end{table}
 
\paragraph{Exact and approximate model selection}
We examined the effectiveness of the
\emph{exact}
and
\emph{approximate}
model selection schemes discussed in
\S~\ref{subsec:app.exact.approximate.model.selection}.
We set 
$C_{\rm min} = 0.01$
and
$C_{\rm max} = 100$.
 The task of
 \emph{exact model selection}
 is to find the best possible regularization parameter  that exactly minimizes the validation error in the continuous range of
 $C \in [C_{\rm min}, C_{\rm max}]$.
 On the other hand, 
 in
 \emph{$\veps$-approximate model selection} scheme, 
 we can find a solution whose validation error is shown to be within $\eps$ from the minimum possible value in that range.
 Starting from
 $C = C_{\rm min}$,
 we gradually increased the regularization parameter
$C$ 
so that the change of the validation errors are within 
$\eps \in \{0, 0.01, 0.05, 0.1\}$.
The results shown in
\figurename~\ref{fig:ExactModelSelection}
and 
Table~\ref{tbl:ExactModelSelection} 
indicate that
the number of models we need to train decreases as $\eps$ increases.

\begin{table}[h]
\begin{center}
 {\small
 \caption{Experimental results on the exact and approximate model selection schemes. The numbers in the table represent how many models were solved in the path.} 
\label{tbl:ExactModelSelection}
 \begin{tabular}{l|rrrr}
Linear LR & $\varepsilon=0.1$ & 0.05 & 0.01 & 0 (exact) \\
\hline
ION & 86 & 205 & 1646 & 13839 \\
BCD & 33 & 66  & 211 & 2654 \\
\hline
\hline
Linear SVM & 0.1 & 0.05 & 0.01 & 0 (exact) \\
\hline
ION & 107 & 230 & 2390 & 17592 \\
BCD & 37 &  77 & 468  & 8817 \\
\hline
\hline
Nonlinear LR & 0.1 & 0.05 & 0.01 & 0 (exact) \\
\hline
BCP & 341 & 633 & 19956 & 19956 \\
PKS & 292 & 523 & 18939 & 18939 \\
\hline
\hline
Nonlinear SVM & 0.1 & 0.05 & 0.01 & 0 (exact) \\
\hline
BCP & 293 & 711 & 9365 & 9365 \\
PKS & 204 & 428 & 19768 & 19768 \\
\hline
 \end{tabular}
}
\end{center}
\end{table}

 \paragraph{Fast LOOCV}
We investigated the efficiency of LOOCV computation in linear LR.
We compared a naive approach (full) and the proposed approach (proposed).
In the naive approach,
$n$ optimization problems in the form of
\eq{eq:LOOCV.computation}
were solved after removing each of the $n$ instances. 
In the proposed approach,
we first computed the model
$w^*_{\rm all}$
by solving the training optimization problem with all the $n$ instances.
Then, 
the lower and the upper bounds of 
$\phi(x_j)^\top w^*_{(-j)}$
were computed 
based on
Theorem~\ref{theo:main}
by using
$w^*_{\rm all}$
as our choice of the suboptimal model.
If the lower bound was larger than 0 or the upper bound was smaller than 0,
we skipped solving the optimization problem 
\eq{eq:LOOCV.computation} 
for that instance.
\figurename~\ref{fig:LOOCV.result}
and
Table~\ref{tab:LOOCV.results}
show the results.
\figurename~\ref{fig:LOOCV.result} indicates that
we could skip solving the optimization problem 
for many instances
especially when the regularization parameter $C$ is small.
From the results in Table~\ref{tab:LOOCV.results},
we could see that the costs of computing the lower and the upper bounds
are negligible compared with the cost of solving optimization problems. 
%
%
%

\begin{table}[h]
 \begin{center}
  \caption{
  The computational time [sec] of LOOCV computation in
  the naive approach (full)
  and
  the proposed approach (proposed)
  for $C \in \{0.01, 1, 100\}$.
  The numbers in the parenthesis are the time taken for computing the
  lower and the upper bounds in Theorem~\ref{theo:main}.
  } 
\label{tab:LOOCV.results}
  {\small
\begin{tabular}{c|c|c|c}
PKS   &$C = 0.01$ & $C = 1$ & $C= 100$ \\ \hline
full & 2.15 & 1.17 & 1.79 \\ 
proposed(bounds) & {\bf 0.41}(0.22) & {\bf 0.37}(0.01) & {\bf 1.44} (0.02) \\ \hline
relative costs & 0.19 & 0.31 & 0.80\\ \hline
  \hline
BCI  &$C = 0.01$ & $C = 1$ & $C= 100$ \\ \hline
full & 6.64& 13.98 & 39.43 \\ 
proposed(bounds) & {\bf 2.58}(0.13) & {\bf 9.39}(0.04) & {\bf 24.39}(0.06) \\ \hline
relative costs& 0.38 & 0.67 & 0.61 \\ \hline
  \hline
BCD   &$C = 0.01$ & $C = 1$ & $C= 100$ \\ \hline
full & 3.94 & 3.57 & 7.49 \\ 
proposed(bounds) & {\bf 0.42}(0.19) & {\bf 0.19}(0.02) & {\bf 1.04}(0.03) \\ \hline
relative costs& 0.10 & 0.053 & 0.13\\ \hline
  \hline
AUS  &$C = 0.01$ & $C = 1$ & $C= 100$ \\ \hline
full & 3.5 & 3.48 & 4.01 \\ 
proposed(bounds) & {\bf 0.24}(0.13) & {\bf 0.6}(0.05) & {\bf 3.32}(0.03) \\ \hline
relative costs& 0.068 & 0.17 & 0.82 \\ \hline
 \hline
G2C   &$C = 0.01$ & $C = 1$ & $C= 100$ \\ \hline
full & 84.04 & 192.94 & 292.26 \\ 
proposed(bounds) & {\bf 11.98}(0.46) & {\bf 62.5}(0.40) & {\bf 127}(0.35) \\ \hline
relative costs& 0.14 & 0.32 & 0.43 \\ \hline
G2N  & $C = 0.01$ & $C = 1$ &$C= 100$ \\ \hline
full & 104.73 & 220.34 & 358.82 \\ 
proposed(bounds) & {\bf 13.87}(0.37) & {\bf 70.7} (0.34) &{\bf 141.71} (0.39) \\ \hline
relative costs& 0.13 & 0.32 & 0.39 \\ \hline
\hline
 SPM   &$C = 0.01$ & $C = 1$ & $C= 100$ \\ \hline
full & 268.68& 794.02 & 2783.59 \\ 
proposed(bounds) &{\bf 3.71}(0.98) & {\bf 180.00}(1.14) & {\bf 1761.4}(1.11) \\ \hline
relative costs& 0.013 & 0.22 & 0.63\\ \hline
  \hline
MGT  &$C = 0.01$ & $C = 1$ & $C= 100$ \\ \hline
full & 1043.7 & 1033.45 & 1064.65 \\ 
proposed(bounds) & {\bf 20.43}(7.16) & {\bf 321.17}(7.34) & {\bf 782.84}(6.19) \\ \hline
relative costs& 0.019 & 0.31 & 0.73\\ \hline
 \end{tabular}
}
 \end{center}
\label{tab:LOOCV_exam}
\end{table}

\paragraph{LR inference by SVM}
Finally,
we present a numerical illustration of LR inferences based on an SVM
solution as discussed in
\S~\ref{subsec:LR.inf.SVM}.
In 
\figurename~\ref{fig:LRbySVM},
the blue circles and the green diamonds represent 
the SVM coefficients 
$w^*_{\rm SVM}$
and 
the optimal LR coefficients 
$w^*_{\rm LR}$,
respectively.
The blue bars indicate the lower and the upper bounds of the optimal LR
coefficients obtained by using the optimal SVM solution as our choice of the suboptimal model.
The top plot is the result obtained by applying
Theorem~\ref{theo:main},
while the bottom plot is the result after considering the intersection of the two balls as described in 
\S~\ref{sec:3}.
The results indicate the advantage of using such an intersection.

\section{Conclusions}
\label{sec:6}
In this paper,
we introduced a novel algorithmic framework for computing the lower and
the upper bounds of the quantities depending on the unknown optimal
solution.
Although we mainly focused on model selection for binary classification problems in this paper, 
our framework can be used in many other machine learning problems.
For example,
we can easily extend our results to LASSO problem
(see Appendix~\ref{app:additional.theory} for details).
As we discussed, 
the choice of the suboptimal model
$\tilde{w}$
is critically important for obtaining useful tight bounds.
An important future work is to develop an algorithm for finding a good
suboptimal model.

\setcounter{figure}{1}

\begin{figure*}[h]
 \begin{center}
  \begin{tabular}{cc}
   \includegraphics[width=0.45\textwidth]{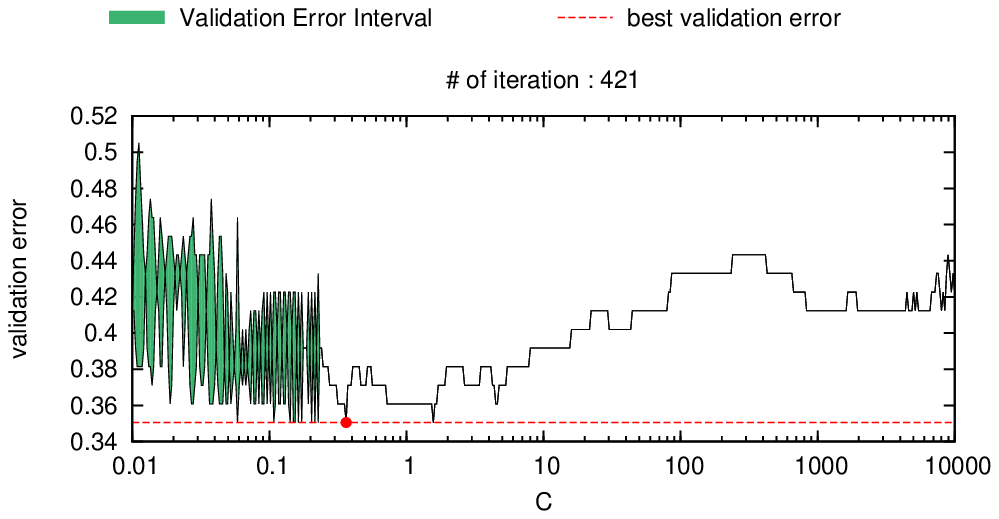} &
   \includegraphics[width=0.45\textwidth]{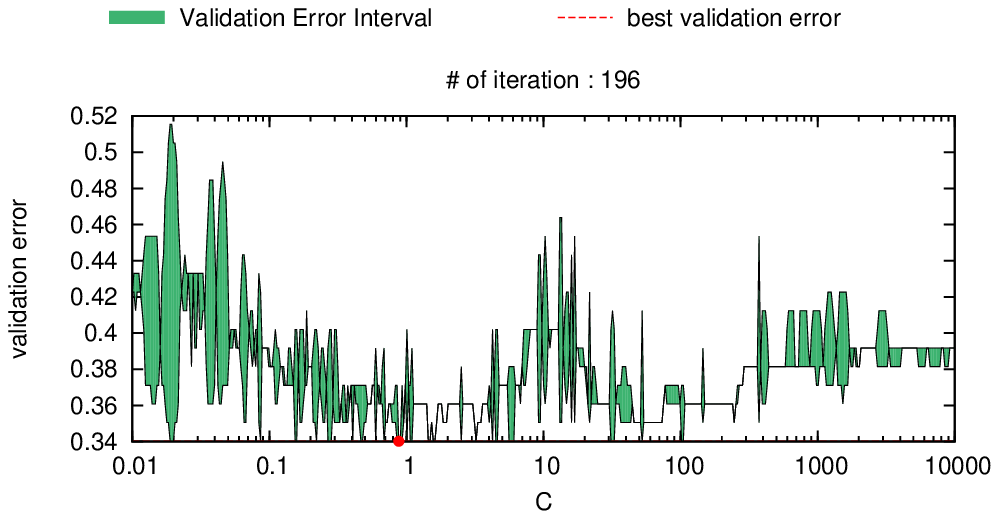} \\
   {\scriptsize{BCP with linear LR}} &
   {\scriptsize{BCP with linear SVM}} \\
   \includegraphics[width=0.45\textwidth]{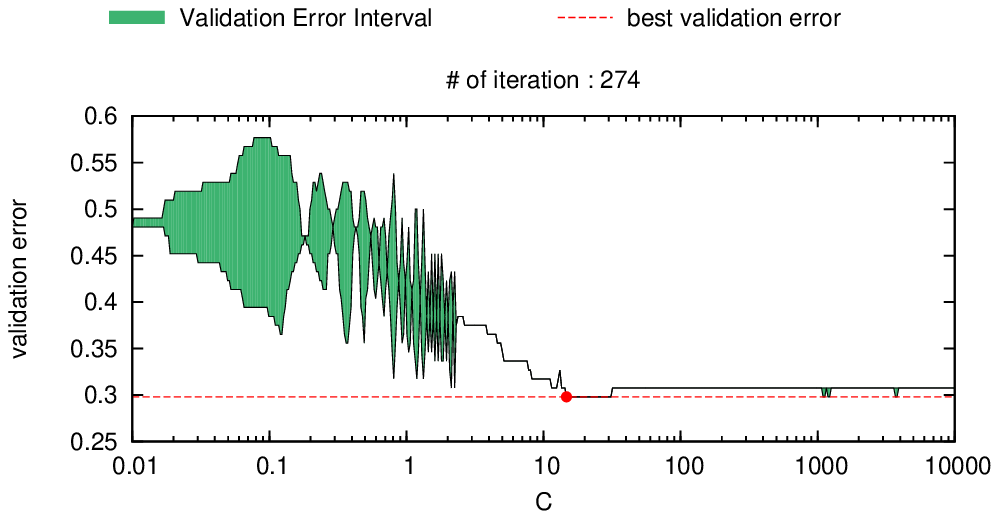} &
   \includegraphics[width=0.45\textwidth]{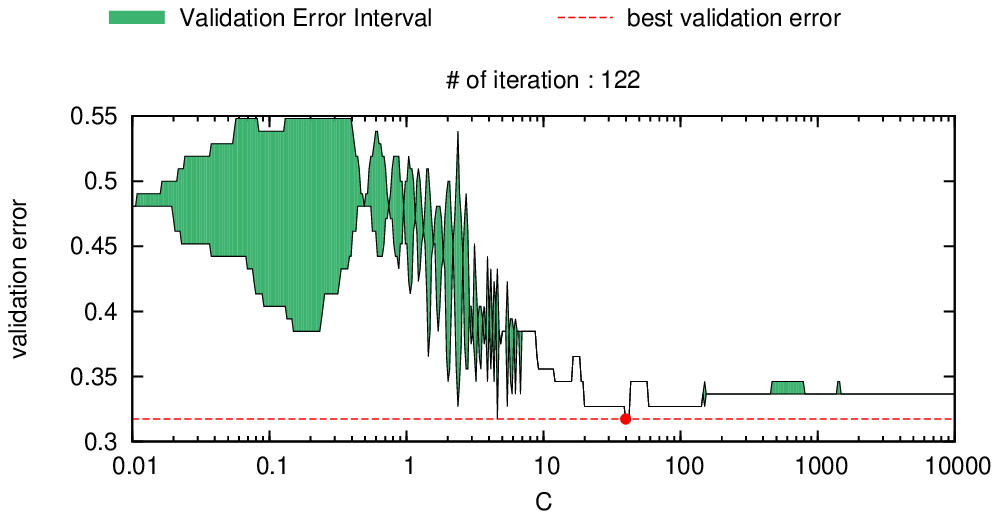} \\
   {\scriptsize{LVD with linear LR}} &
   {\scriptsize{LVD with linear SVM}} \\
   \includegraphics[width=0.45\textwidth]{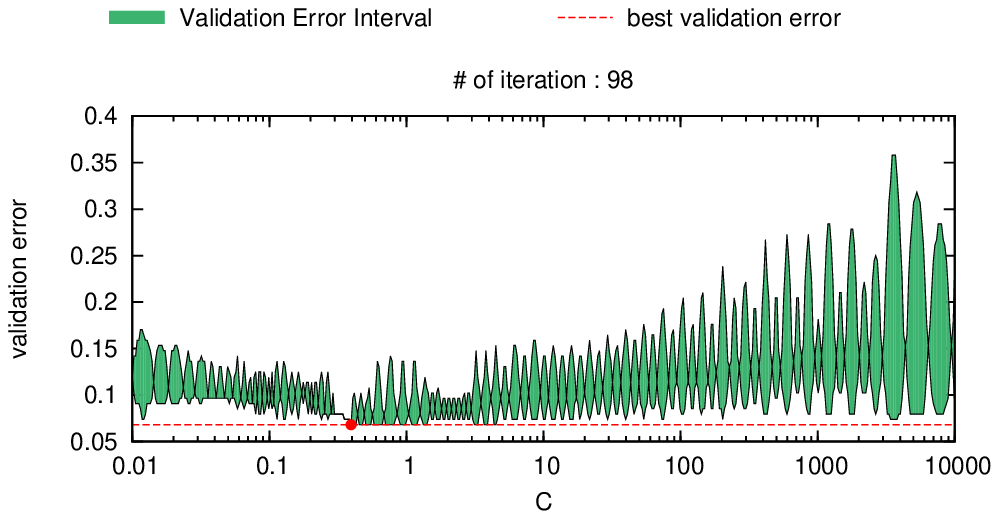} &
   \includegraphics[width=0.45\textwidth]{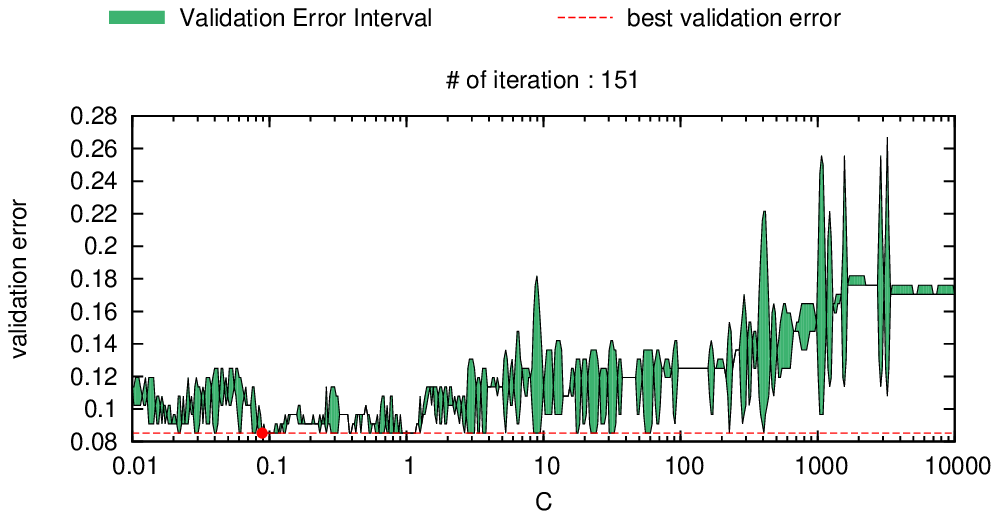} \\
   {\scriptsize{ION with linear LR}} &
   {\scriptsize{ION with linear SVM}} \\
   \includegraphics[width=0.45\textwidth]{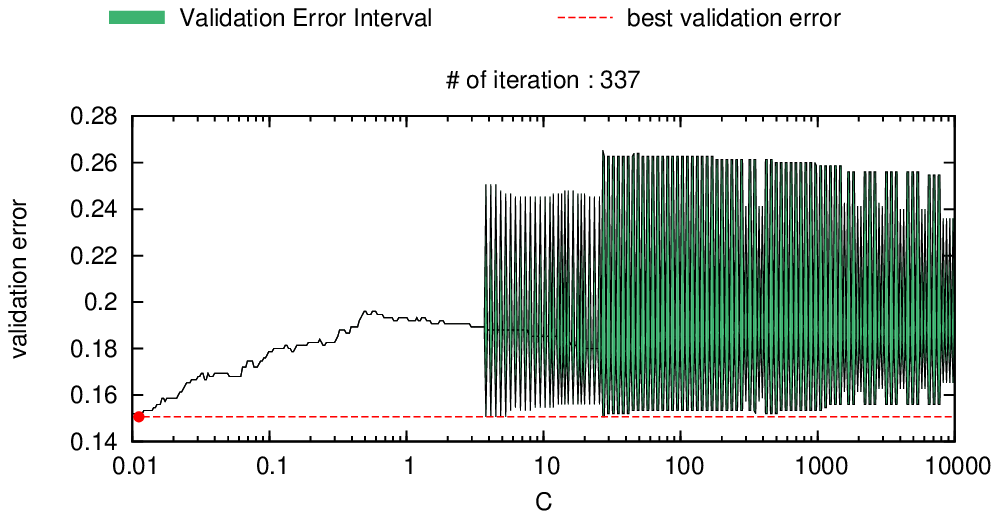} &
   \includegraphics[width=0.45\textwidth]{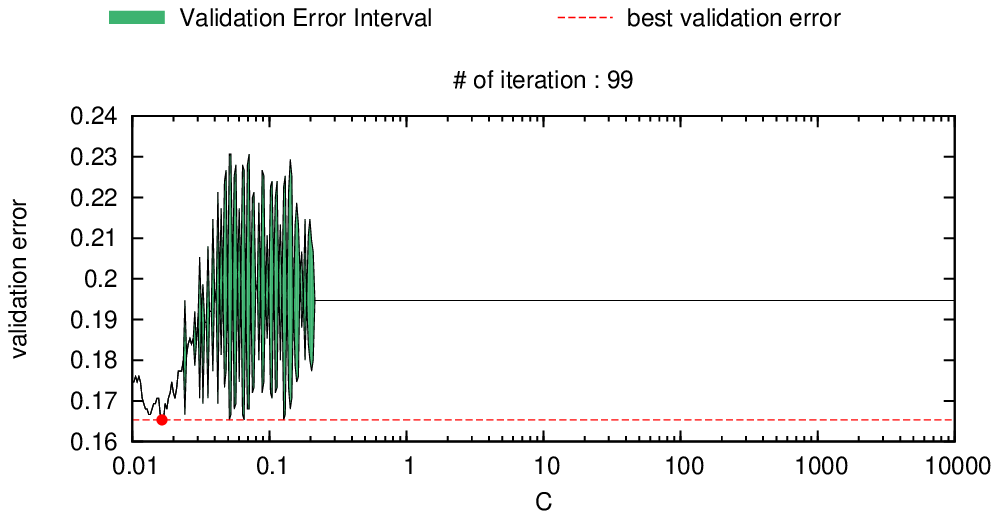} \\
   {\scriptsize{G2C with linear LR}} &
   {\scriptsize{G2C with linear SVM}} \\
  \end{tabular}
 \caption{The sequence of validation error bounds after the final step of
  the efficient model selection processes. Although the validation errors with several 
  regularization parameters are still unknown, we can guarantee
  that the current smallest solution (red point) is the best one.}
 \label{fig:exp.ems}
 \end{center}
\end{figure*}

\setcounter{figure}{1}

\begin{figure*}[h]
 \begin{center}
  \begin{tabular}{cc}
   \includegraphics[width=0.45\textwidth]{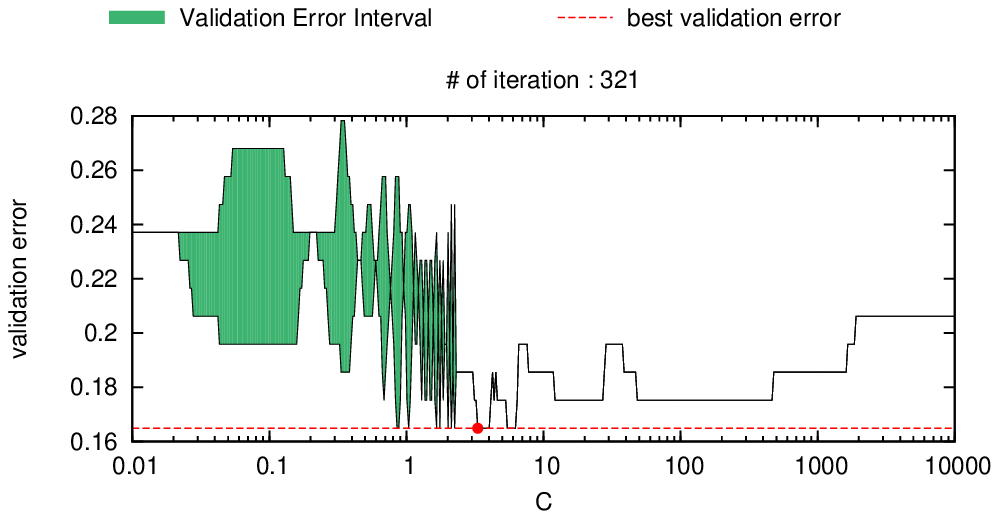} &
   \includegraphics[width=0.45\textwidth]{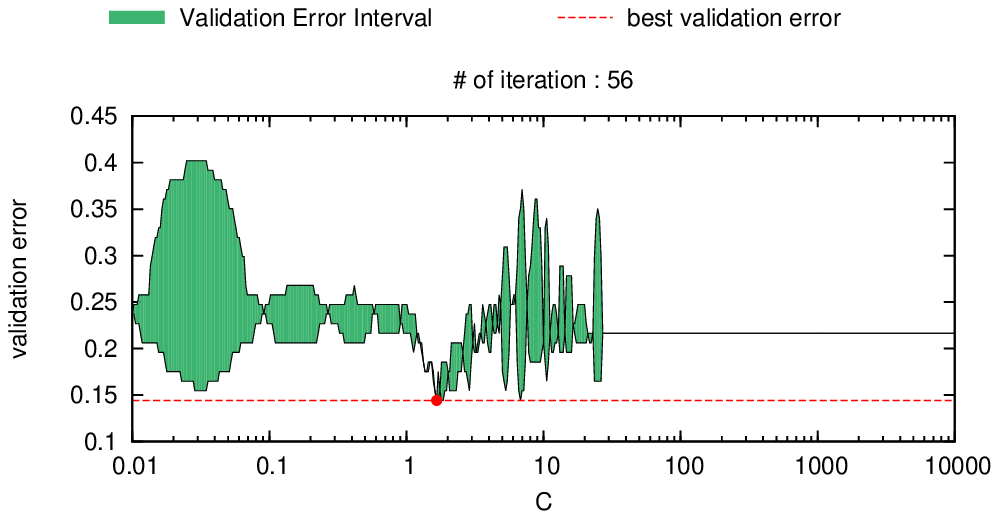} \\
   {\scriptsize{BCP with nonlinear LR}} &
   {\scriptsize{BCP with nonlinear SVM}} \\
   \includegraphics[width=0.45\textwidth]{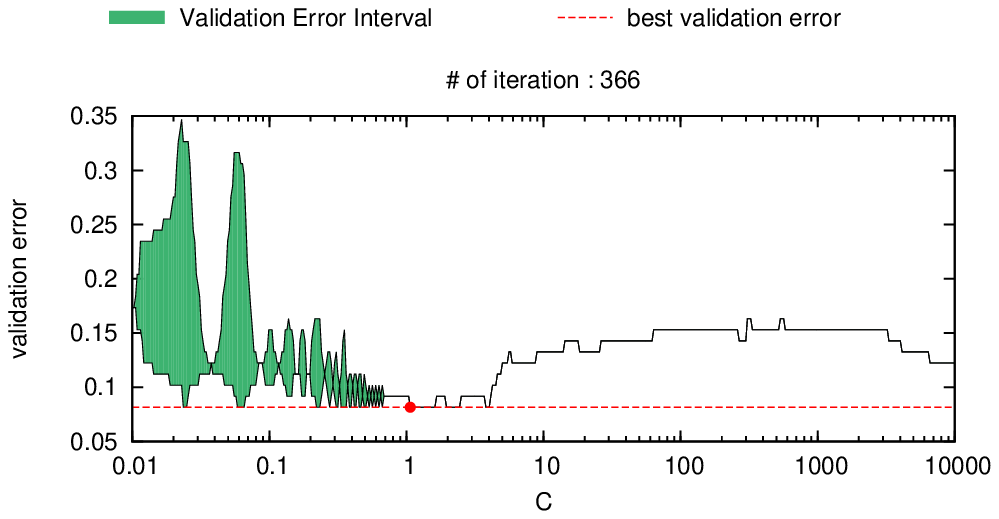} &
   \includegraphics[width=0.45\textwidth]{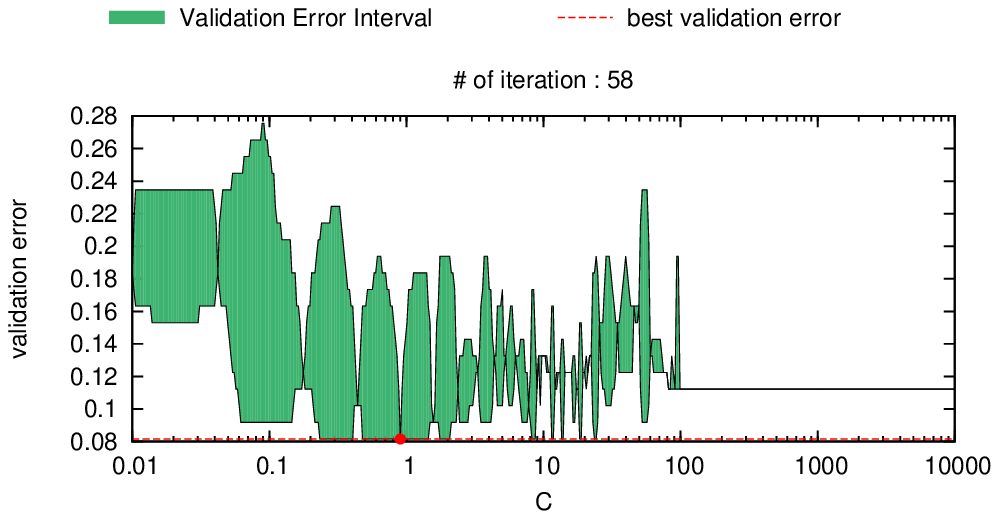} \\
   {\scriptsize{PKS with nonlinear LR}} &
   {\scriptsize{PKS with nonlinear SVM}} \\
   \includegraphics[width=0.45\textwidth]{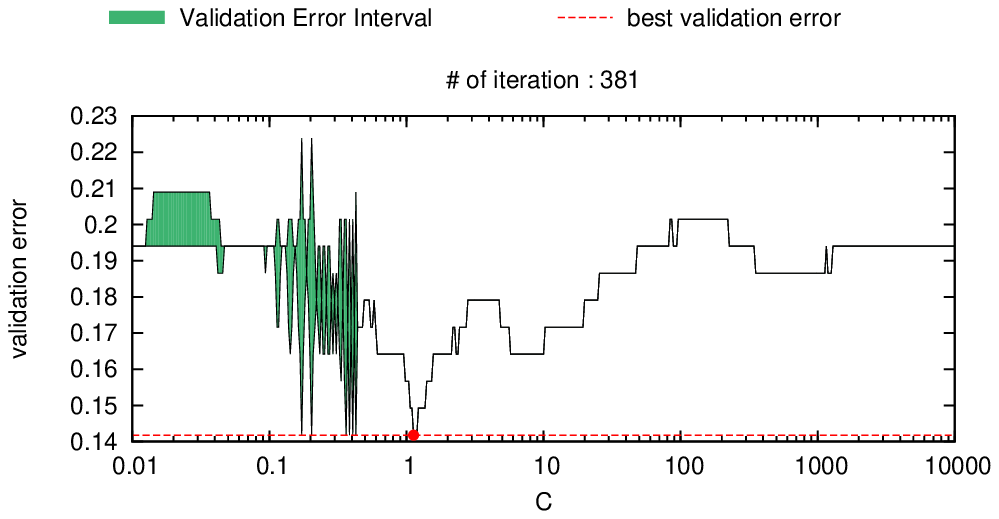} &
   \includegraphics[width=0.45\textwidth]{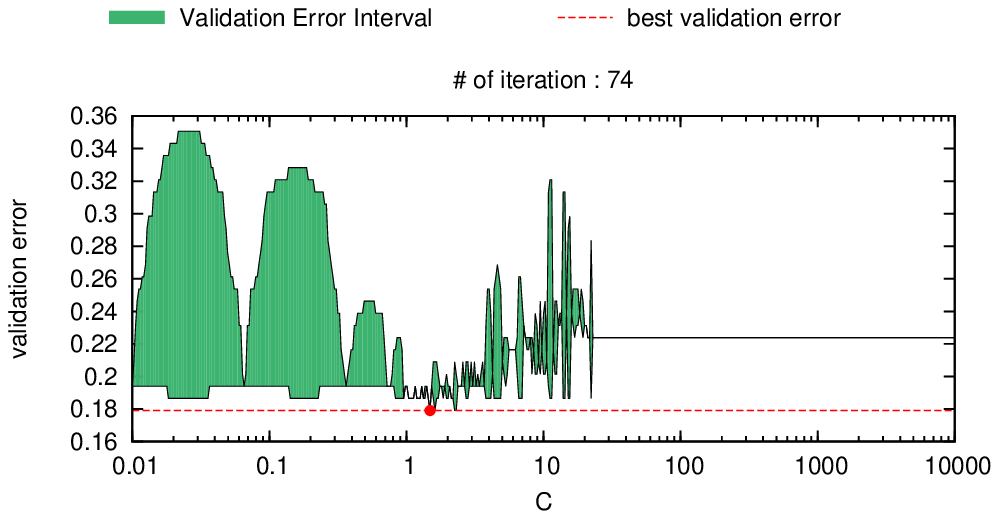} \\
   {\scriptsize{SPH with nonlinear LR}} &
   {\scriptsize{SPH with nonlinear SVM}} \\
   \includegraphics[width=0.45\textwidth]{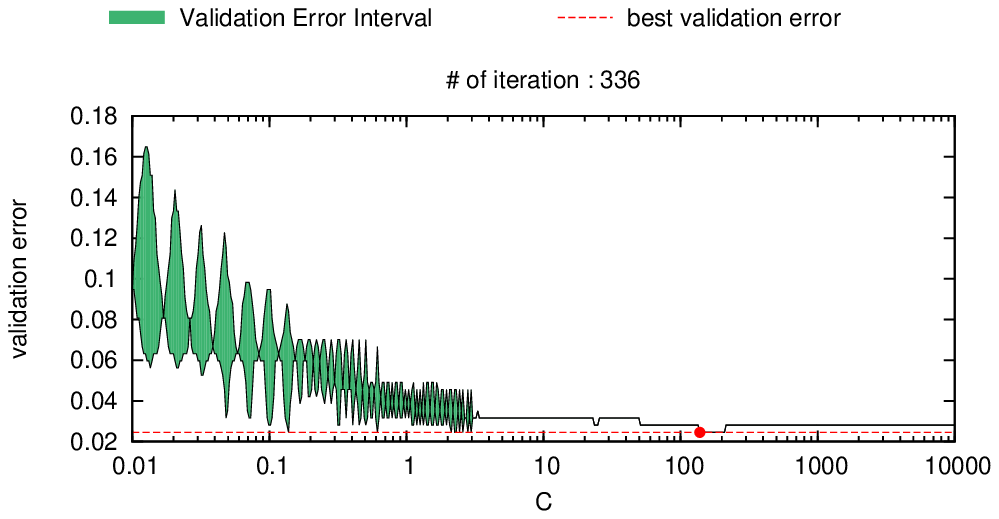} &
   \includegraphics[width=0.45\textwidth]{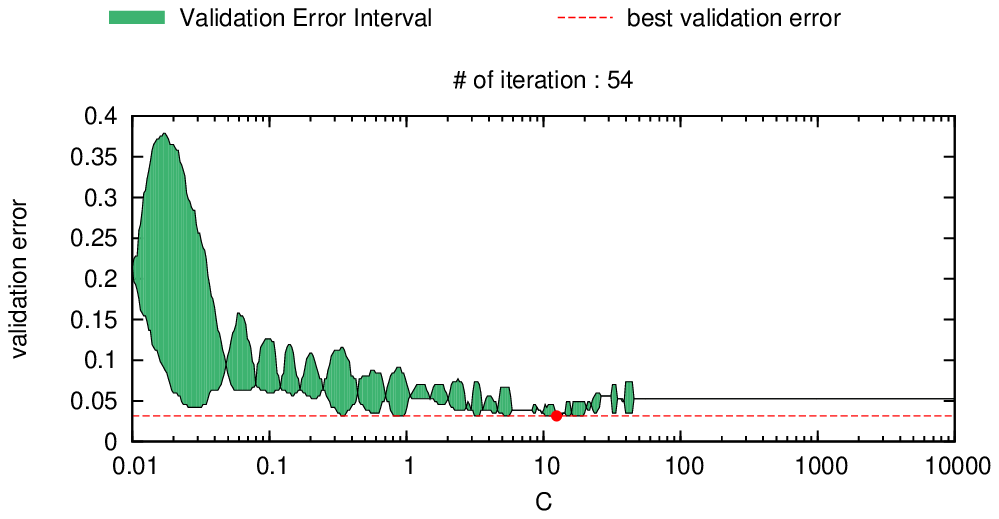} \\
   {\scriptsize{BCD with nonlinear LR}} &
   {\scriptsize{BCD with nonlinear SVM}} 
  \end{tabular}
 \caption{Continued.}
 \label{fig:exp.ems}
 \end{center}
\end{figure*}

\setcounter{figure}{2}
\vspace{-30mm}

\begin{figure*}[htbp]
 \begin{center}
  \begin{tabular}{cc}
   \includegraphics[width=0.45\textwidth]{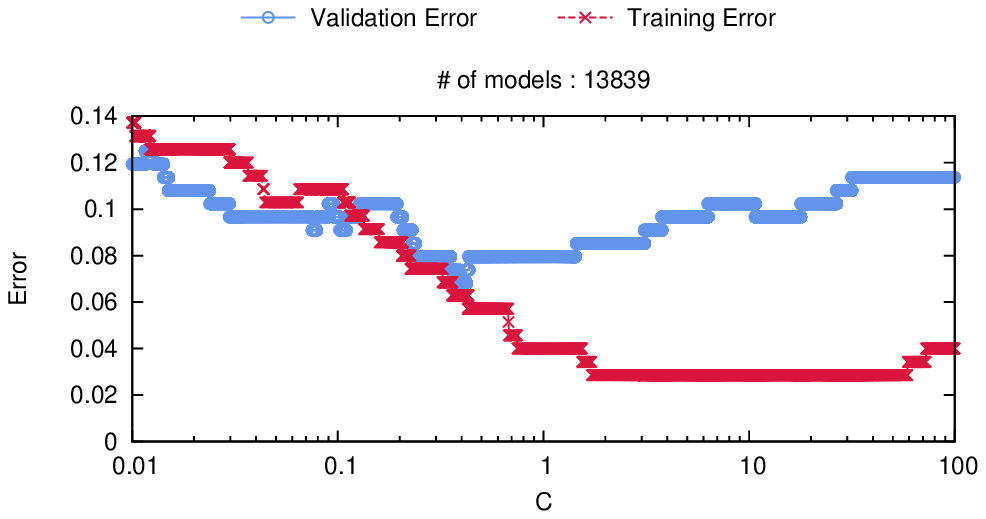} &
   \includegraphics[width=0.45\textwidth]{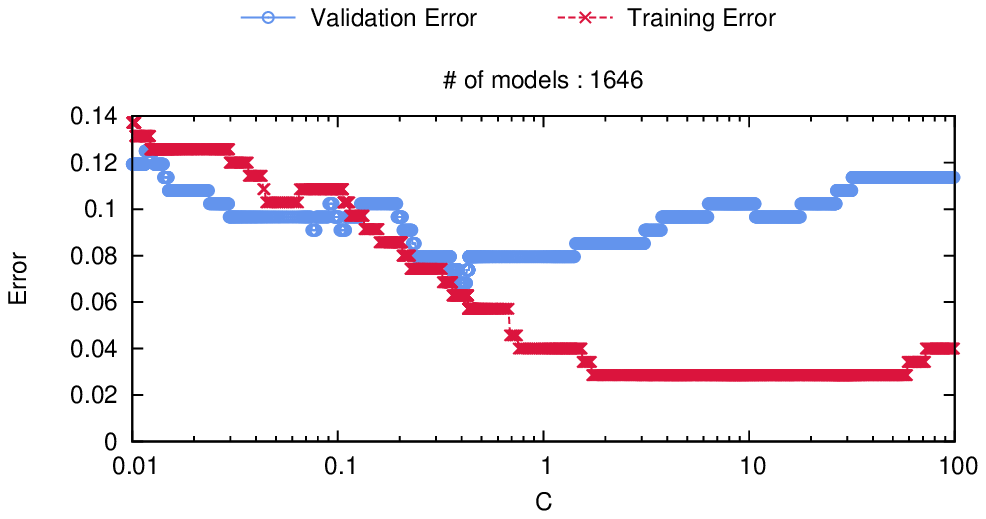} \\
   {\scriptsize (a1) linear LR with $\veps = 0$ on ION} &
   {\scriptsize (a2) linear LR with $\veps = 0.01$ on ION} \\
   \includegraphics[width=0.45\textwidth]{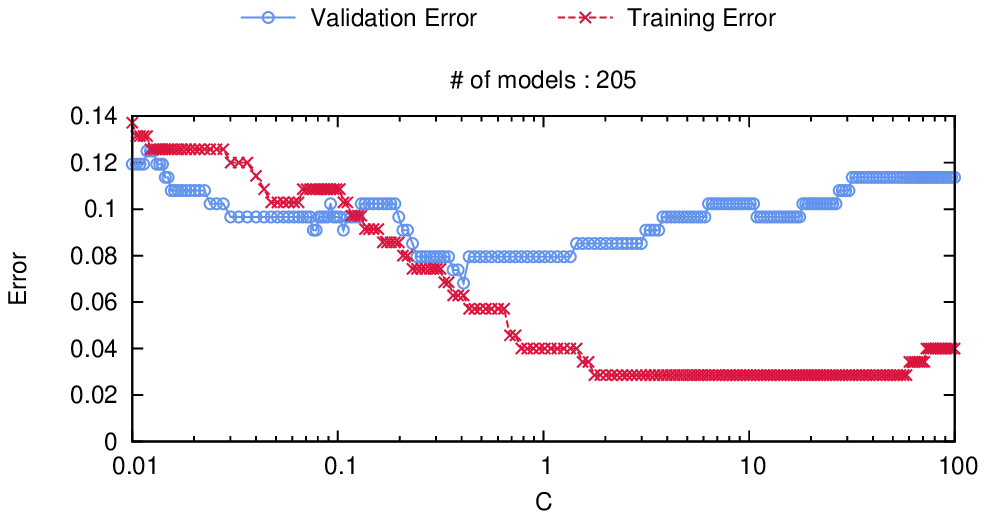} &
   \includegraphics[width=0.45\textwidth]{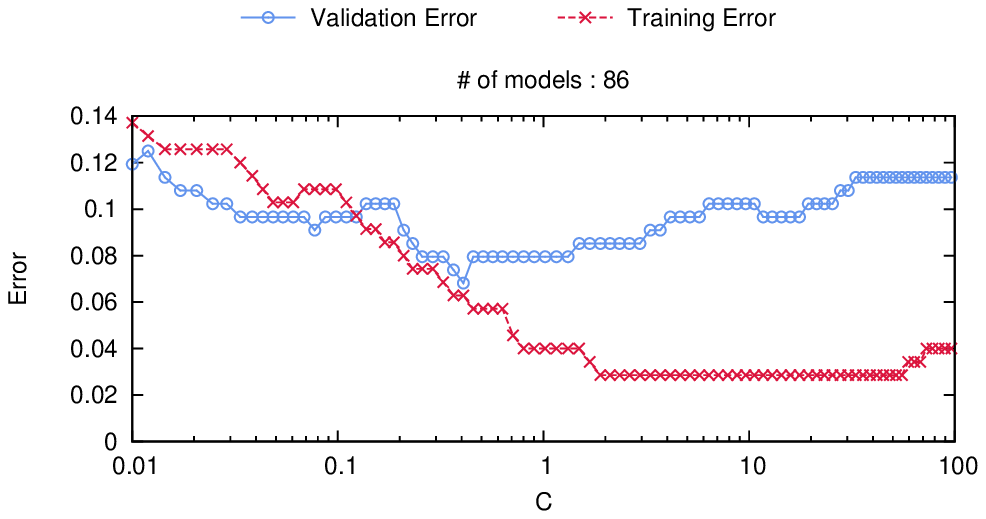} \\
   {\scriptsize (a3) linear LR with $\veps = 0.05$ on ION} &
   {\scriptsize (a4) linear LR with $\veps = 0.1$ on ION} \\
   \includegraphics[width=0.45\textwidth]{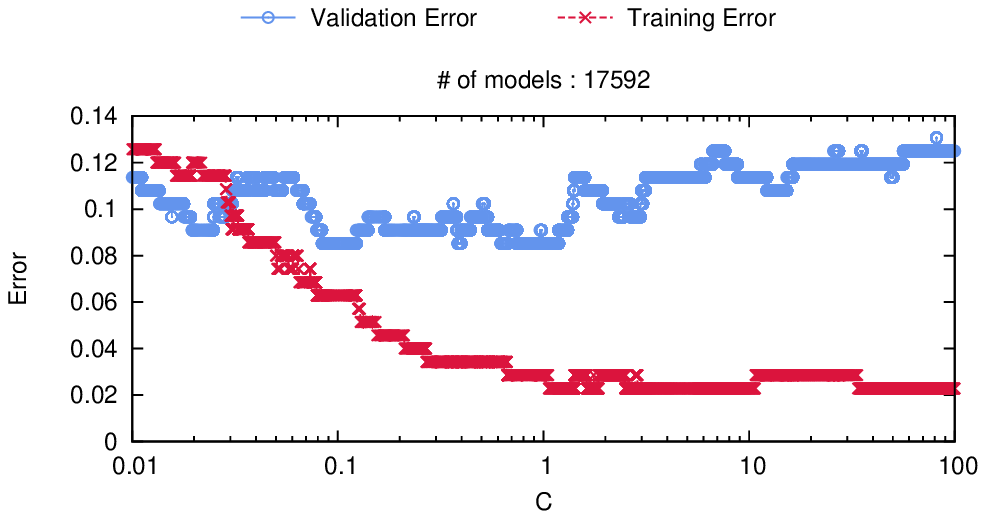} &
   \includegraphics[width=0.45\textwidth]{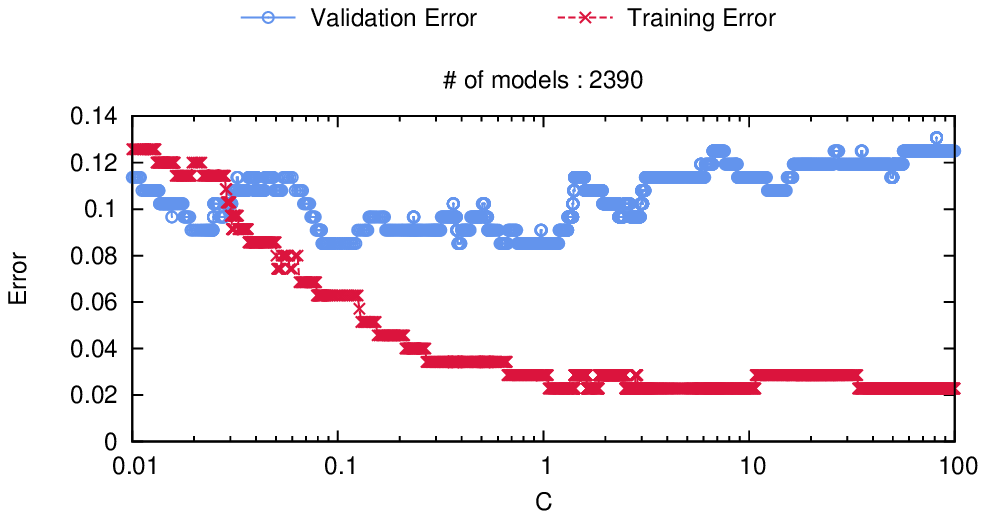} \\
   {\scriptsize (a5) linear SVM with $\veps = 0$ on ION} &
   {\scriptsize (a6) linear SVM with $\veps = 0.01$ on ION} \\
   \includegraphics[width=0.45\textwidth]{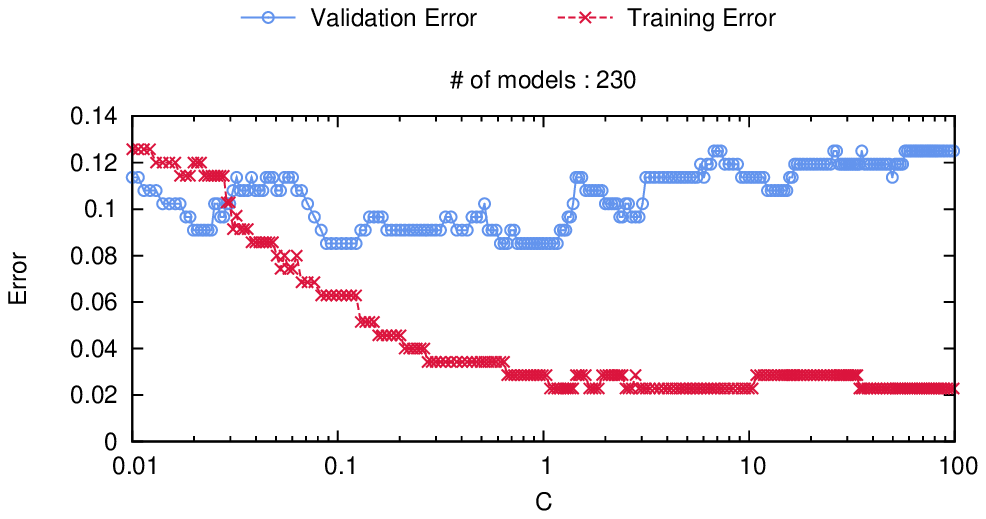} &
   \includegraphics[width=0.45\textwidth]{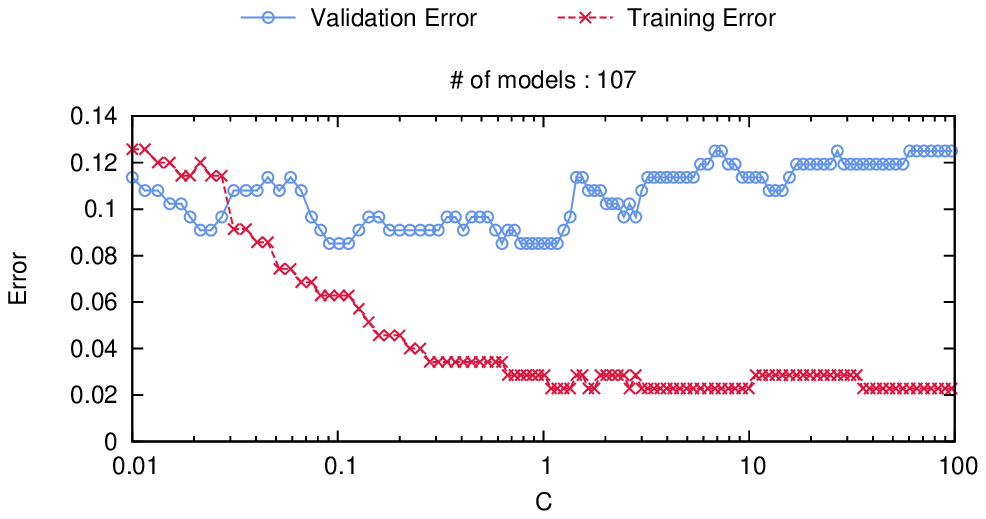} \\
   {\scriptsize (a7) linear SVM with $\veps = 0.05$ on ION} &
   {\scriptsize (a8) linear SVM with $\veps = 0.1$ on ION} \\
  \end{tabular}
  \caption{Exact and approximate model selection results.
  The validation error of the solution is
  shown to be within $\veps$ from the smallest possible value in the
  continuous range of $C \in [C_{\rm min}, C_{\rm max}]$.
  }
  \label{fig:ExactModelSelection}
 \end{center}
\end{figure*}

\setcounter{figure}{2}

\begin{figure*}[htbp]
 \begin{center}
  \begin{tabular}{cc}
   \includegraphics[width=0.45\textwidth]{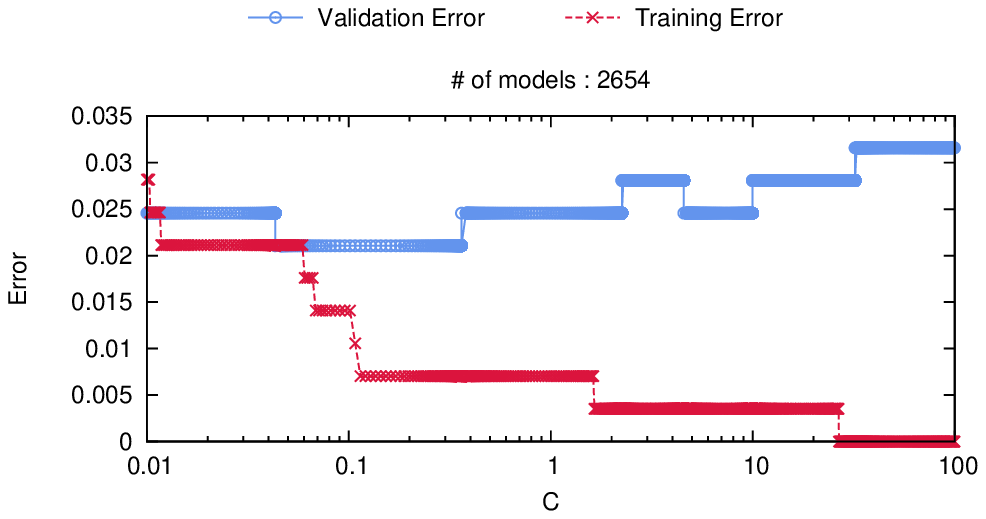} &
   \includegraphics[width=0.45\textwidth]{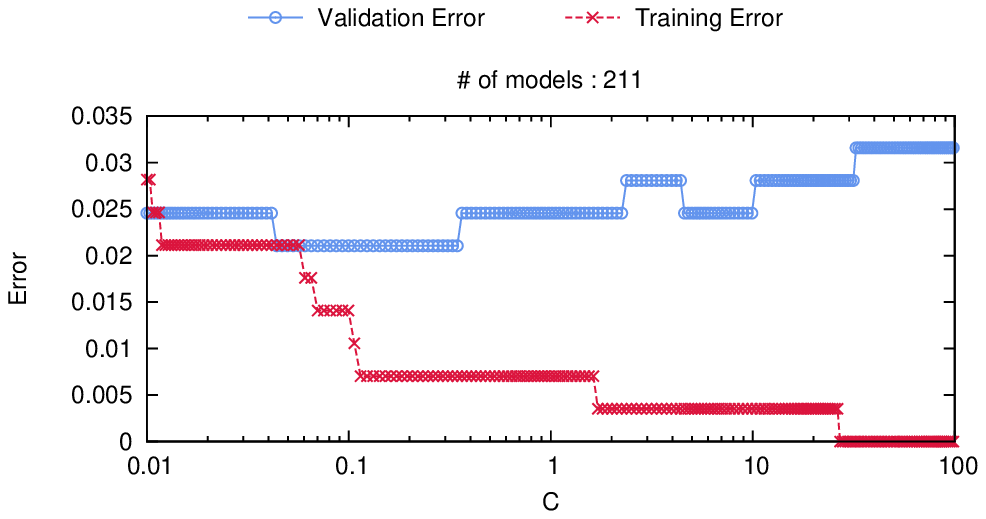} \\
   {\scriptsize (b1) linear LR with $\veps = 0$ on BCD} &
   {\scriptsize (b2) linear LR with $\veps = 0.01$ on BCD} \\
   \includegraphics[width=0.45\textwidth]{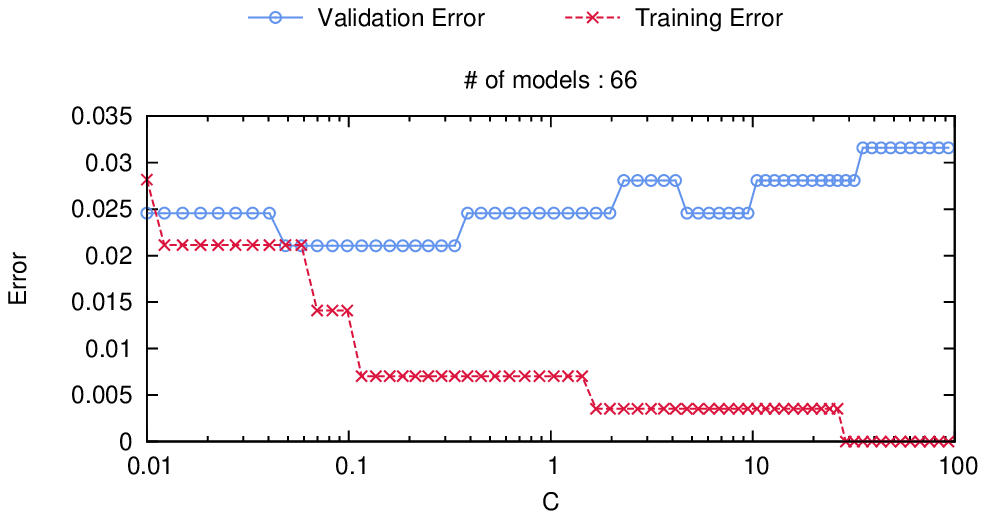} &
   \includegraphics[width=0.45\textwidth]{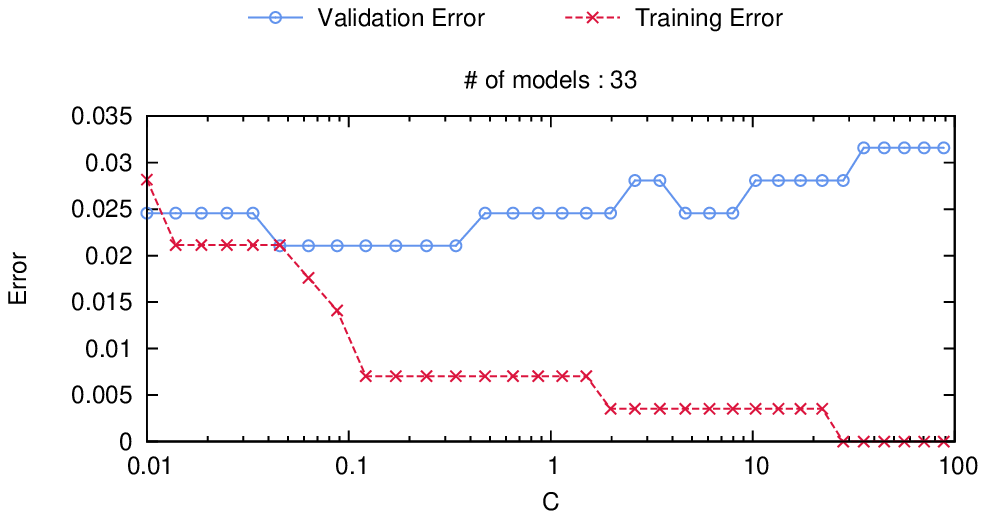} \\
   {\scriptsize (b3) linear LR with $\veps = 0.05$ on BCD} &
   {\scriptsize (b4) linear LR with $\veps = 0.1$ on BCD} \\
   \includegraphics[width=0.45\textwidth]{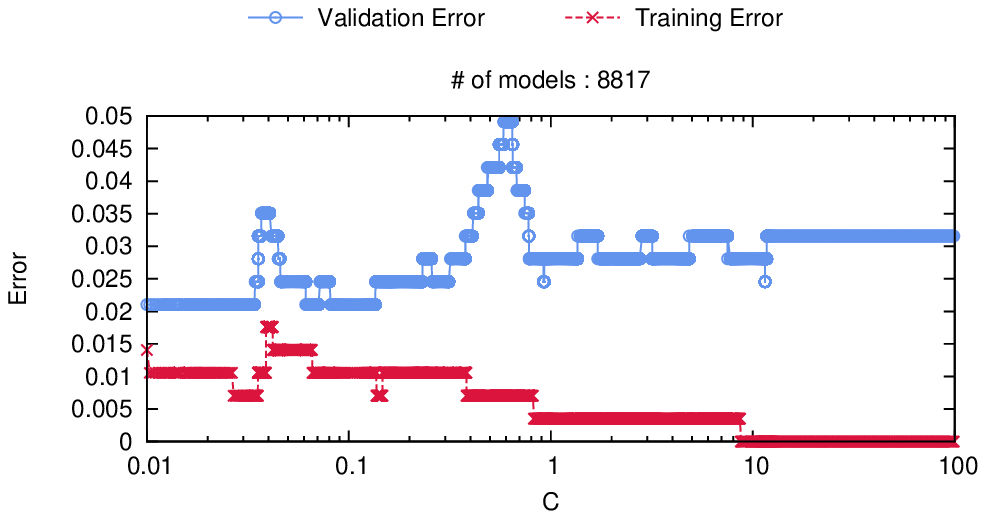} &
   \includegraphics[width=0.45\textwidth]{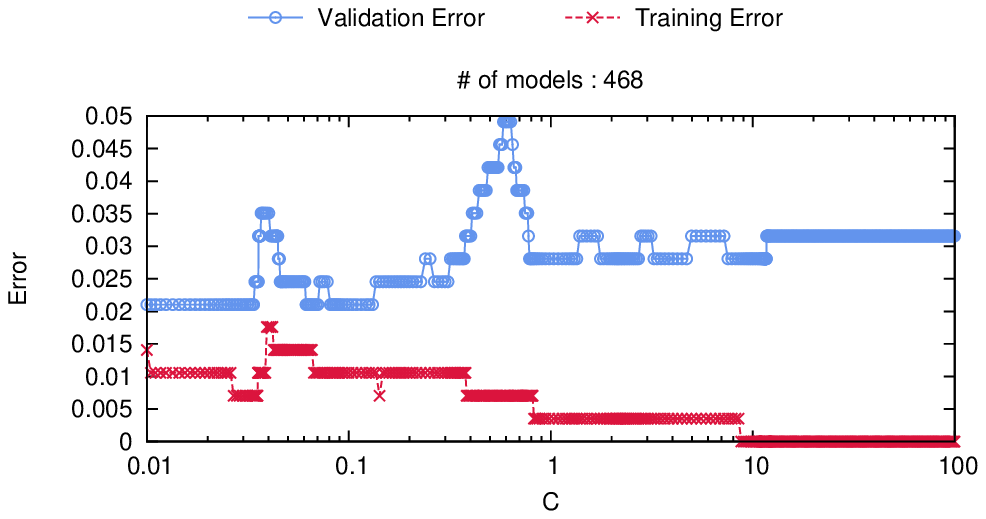} \\
   {\scriptsize (b5) linear SVM with $\veps = 0$ on BCD} &
   {\scriptsize (b6) linear SVM with $\veps = 0.01$ on BCD} \\
   \includegraphics[width=0.45\textwidth]{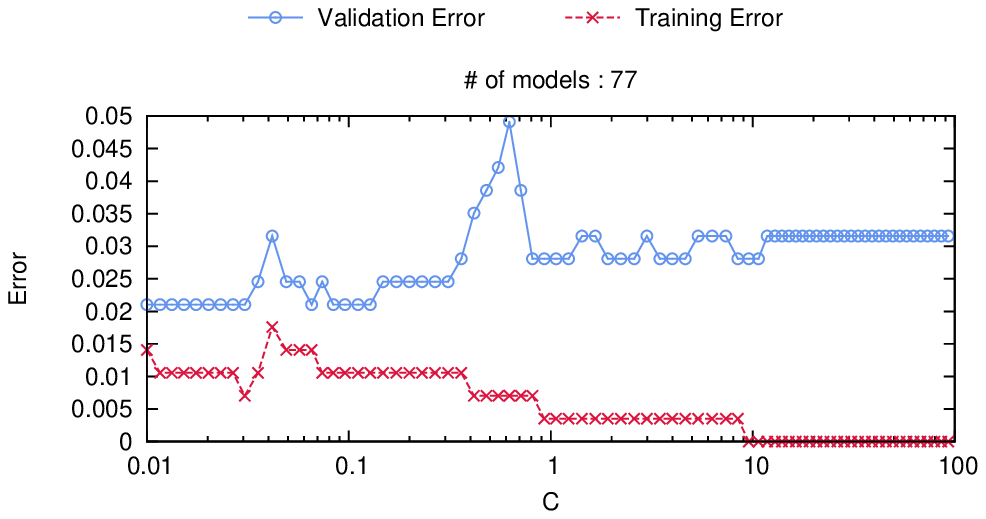} &
   \includegraphics[width=0.45\textwidth]{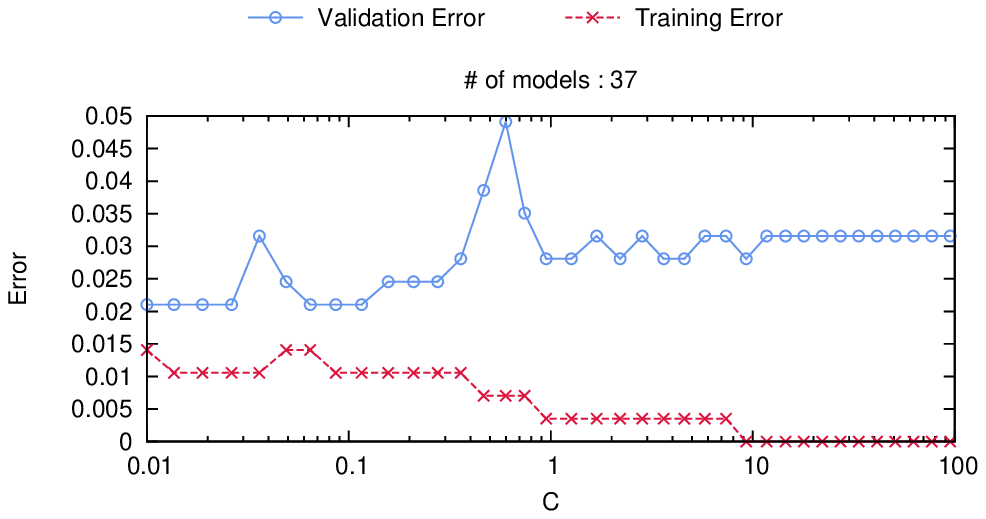} \\
   {\scriptsize (b7) linear SVM with $\veps = 0.05$ on BCD} &
   {\scriptsize (b8) linear SVM with $\veps = 0.1$ on BCD} \\
  \end{tabular}
  \caption{Continued.}
 \end{center}
\end{figure*}

\setcounter{figure}{2}

\begin{figure*}[htbp]
 \begin{center}
  \begin{tabular}{cc}
   \includegraphics[width=0.45\textwidth]{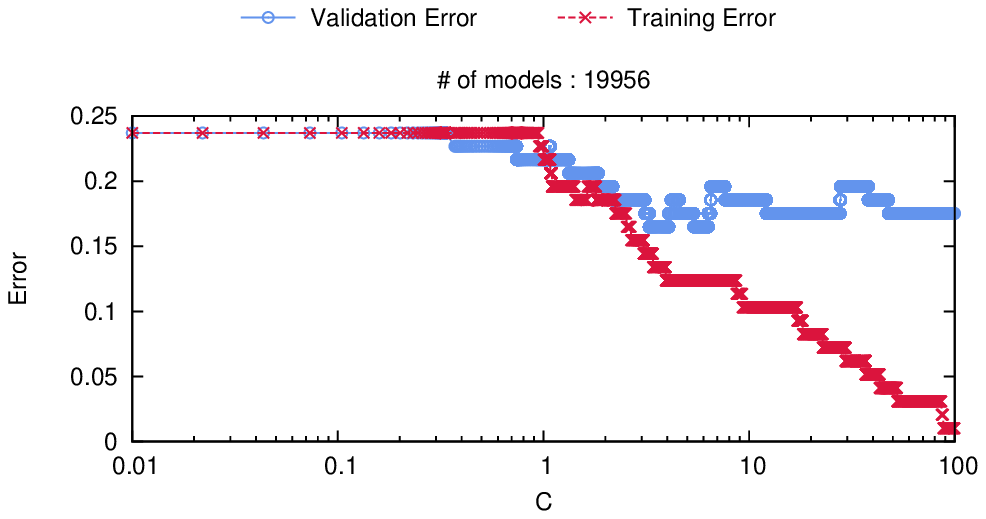} &
   \includegraphics[width=0.45\textwidth]{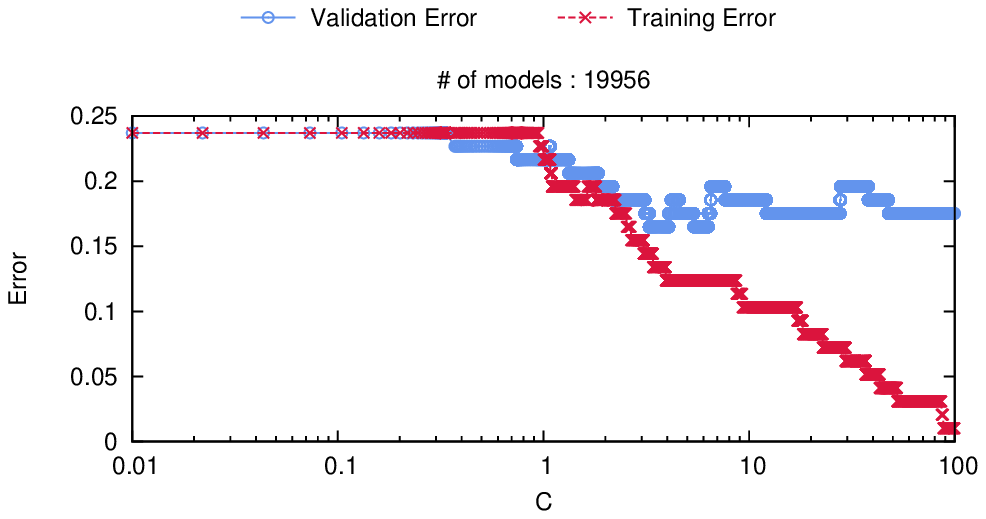} \\
   {\scriptsize (c1) nonlinear LR with $\veps = 0$ on BCP} &
   {\scriptsize (c2) nonlinear LR with $\veps = 0.01$ on BCP} \\
   \includegraphics[width=0.45\textwidth]{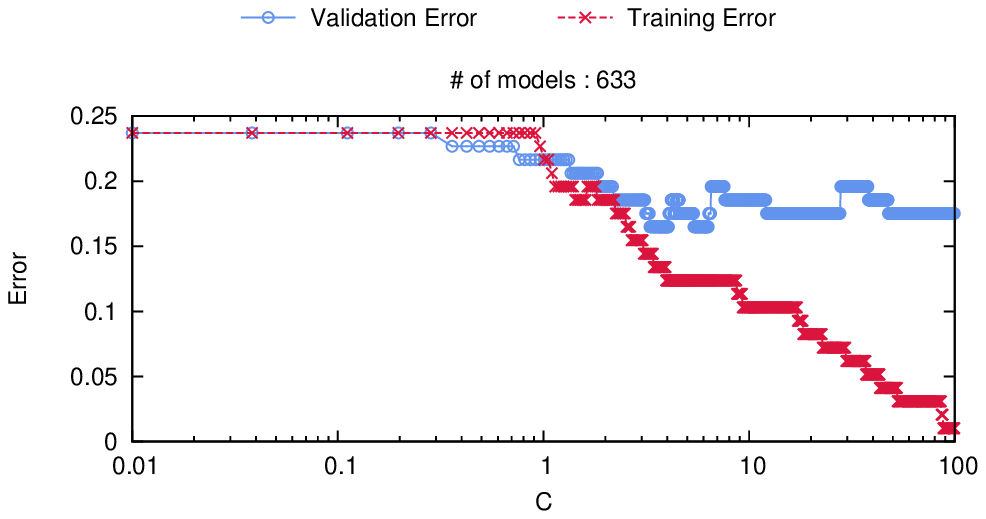} &
   \includegraphics[width=0.45\textwidth]{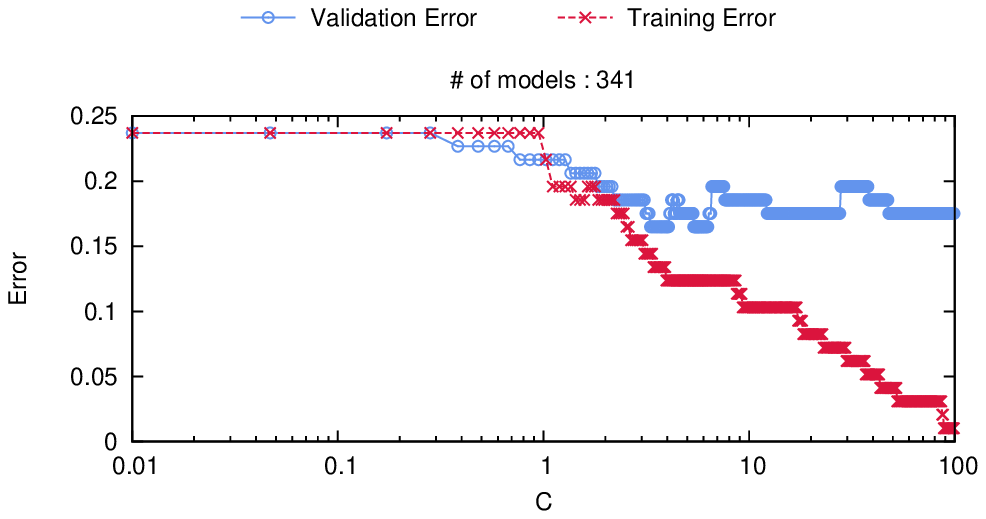} \\
   {\scriptsize (c3) nonlinear LR with $\veps = 0.05$ on BCP} &
   {\scriptsize (c4) nonlinear LR with $\veps = 0.1$ on BCP} \\
   \includegraphics[width=0.45\textwidth]{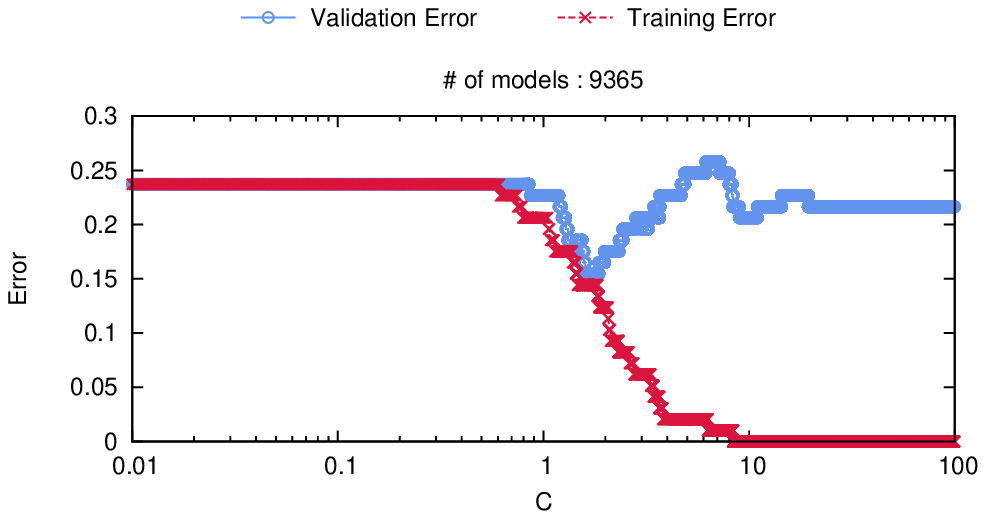} &
   \includegraphics[width=0.45\textwidth]{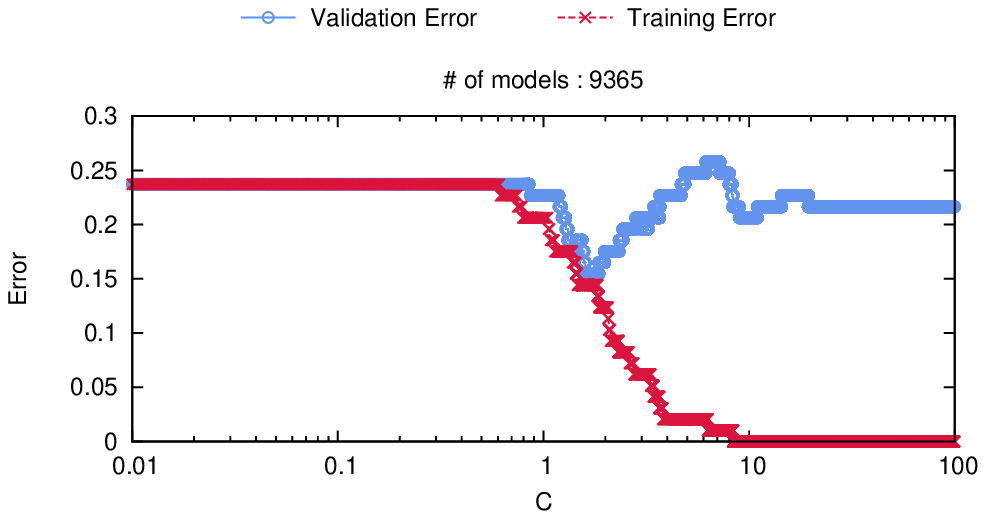} \\
   {\scriptsize (c5) nonlinear SVM with $\veps = 0$ on BCP} &
   {\scriptsize (c6) nonlinear SVM with $\veps = 0.01$ on BCP} \\
   \includegraphics[width=0.45\textwidth]{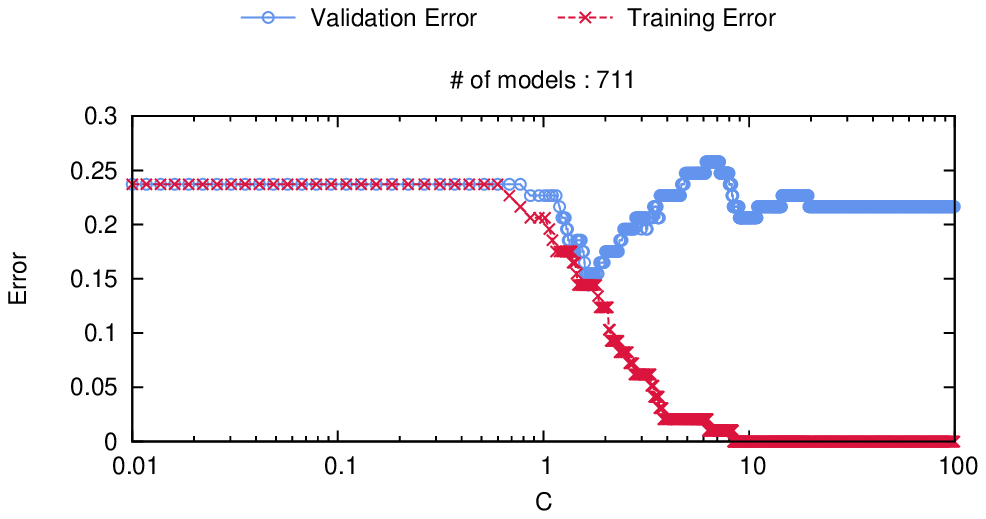} &
   \includegraphics[width=0.45\textwidth]{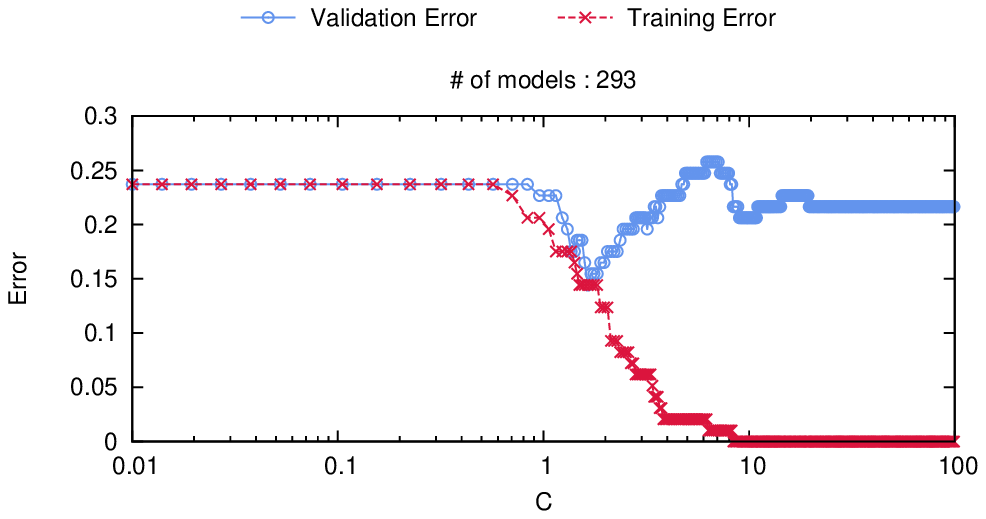} \\
   {\scriptsize (c7) nonlinear SVM with $\veps = 0.05$ on BCP} &
   {\scriptsize (c8) nonlinear SVM with $\veps = 0.1$ on BCP} \\
  \end{tabular}
  \caption{Continued.}
 \end{center}
\end{figure*}

\setcounter{figure}{2}

\begin{figure*}[htbp]
 \begin{center}
  \begin{tabular}{cc}
   \includegraphics[width=0.45\textwidth]{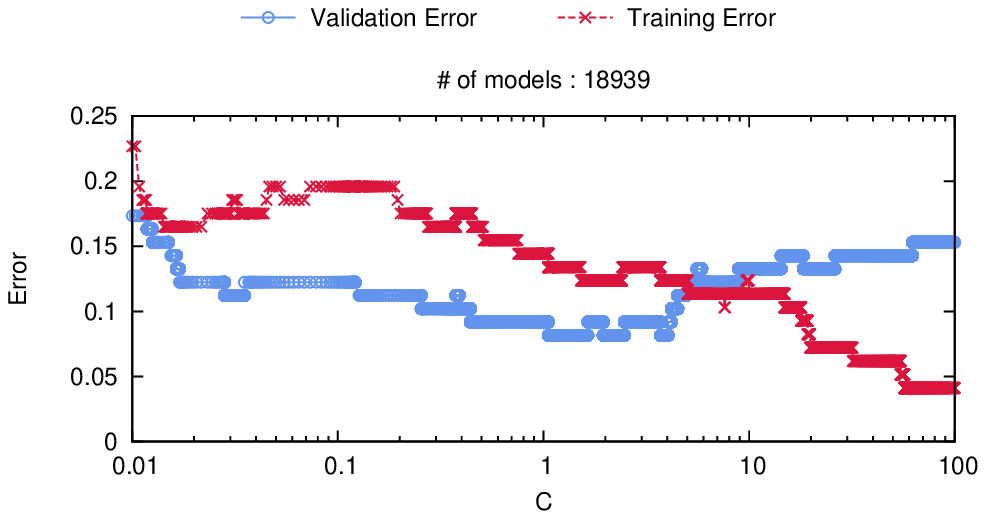} &
   \includegraphics[width=0.45\textwidth]{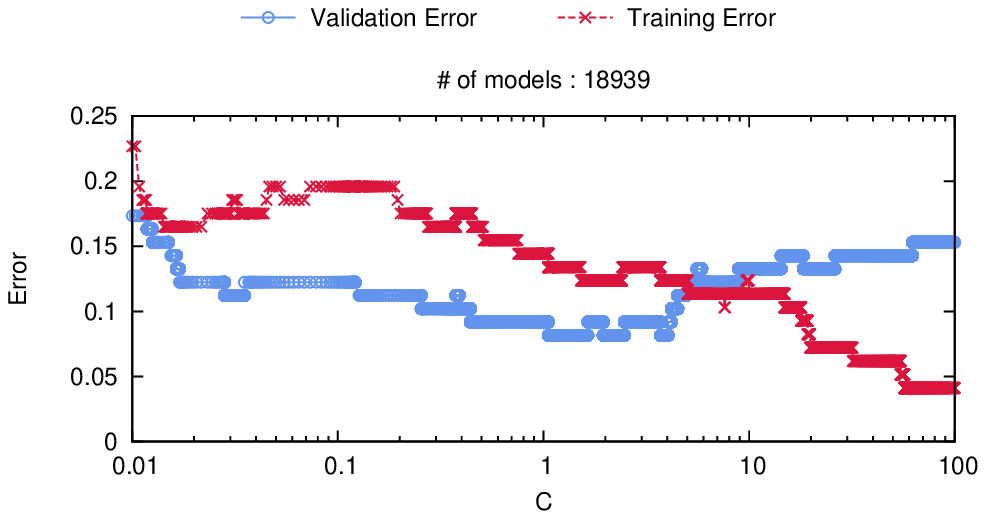} \\
   {\scriptsize (d1) nonlinear LR with $\veps = 0$ on BCP} &
   {\scriptsize (d2) nonlinear LR with $\veps = 0.01$ on BCP} \\
   \includegraphics[width=0.45\textwidth]{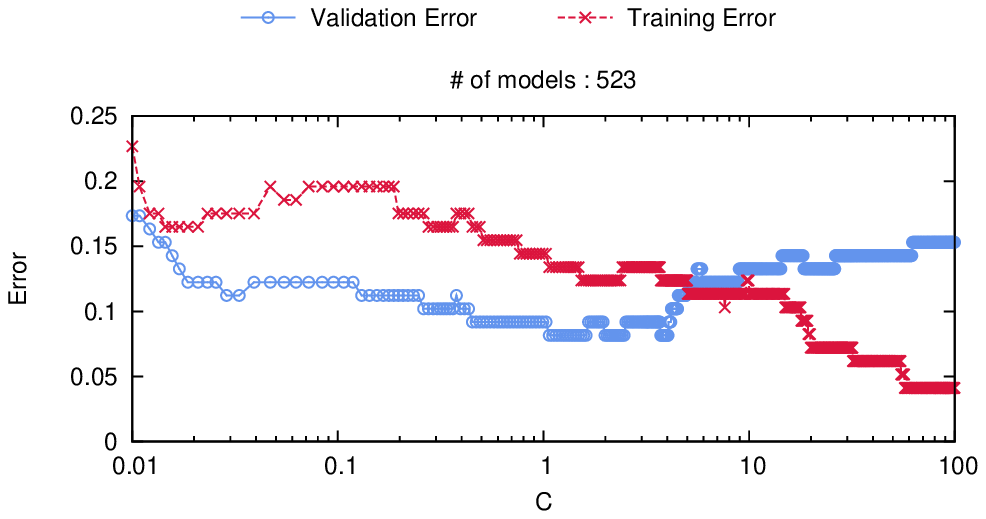} &
   \includegraphics[width=0.45\textwidth]{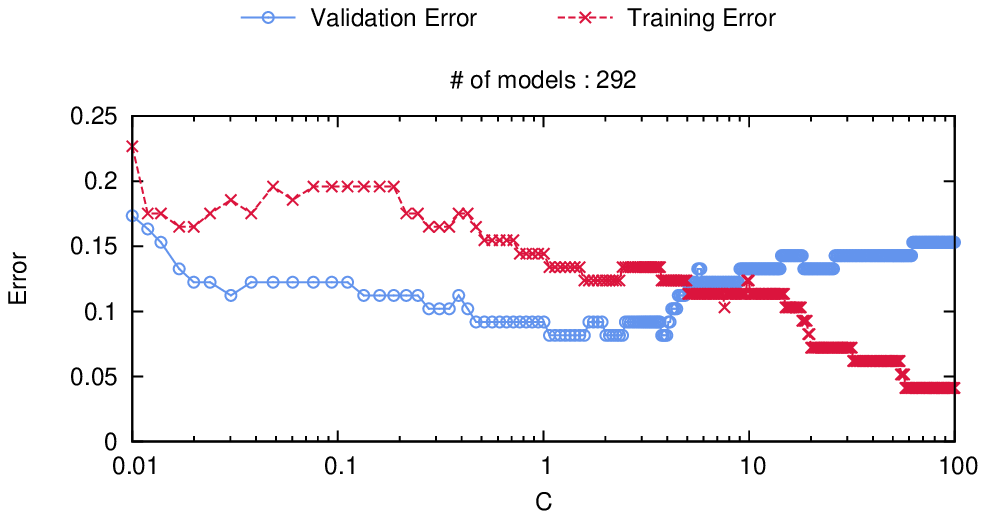} \\
   {\scriptsize (d3) nonlinear LR with $\veps = 0.05$ on BCP} &
   {\scriptsize (d4) nonlinear LR with $\veps = 0.1$ on BCP} \\
   \includegraphics[width=0.45\textwidth]{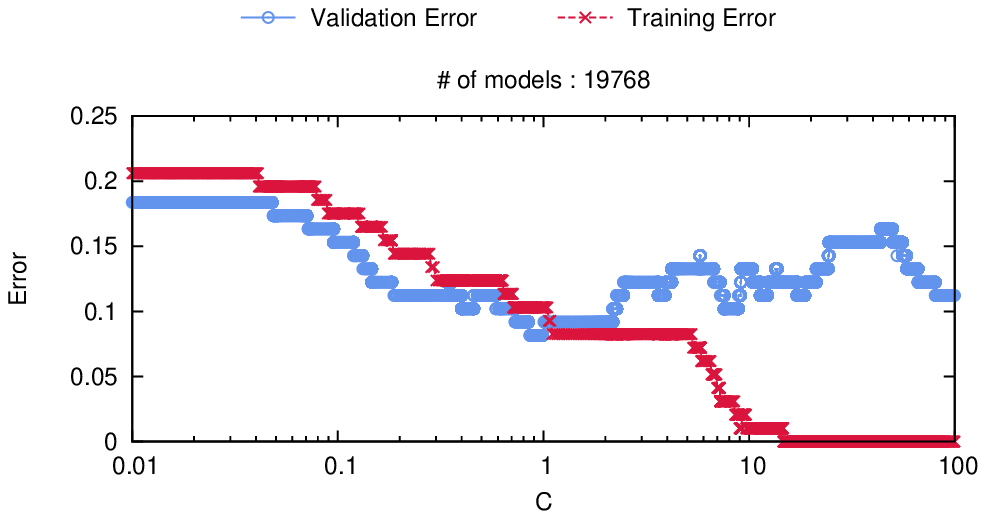} &
   \includegraphics[width=0.45\textwidth]{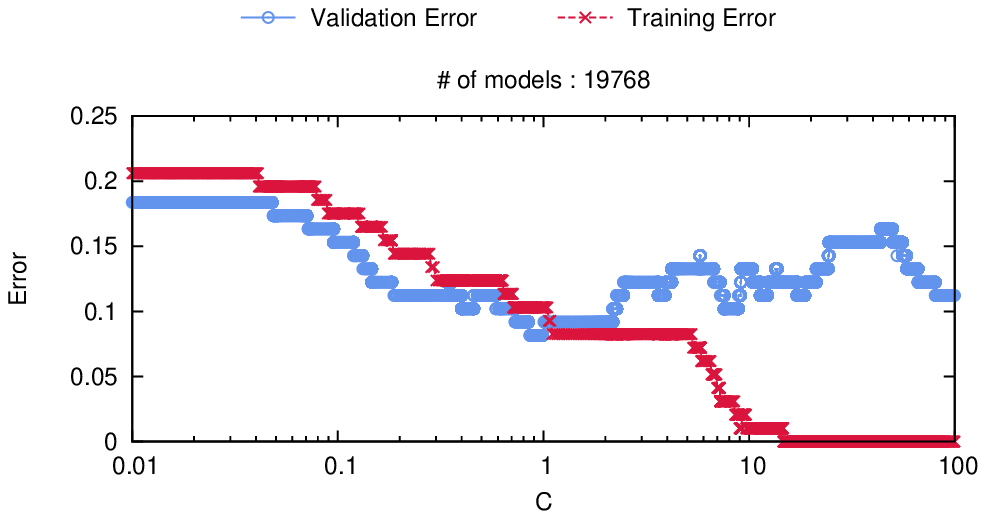} \\
   {\scriptsize (d5) nonlinear SVM with $\veps = 0$ on BCP} &
   {\scriptsize (d6) nonlinear SVM with $\veps = 0.01$ on BCP} \\
   \includegraphics[width=0.45\textwidth]{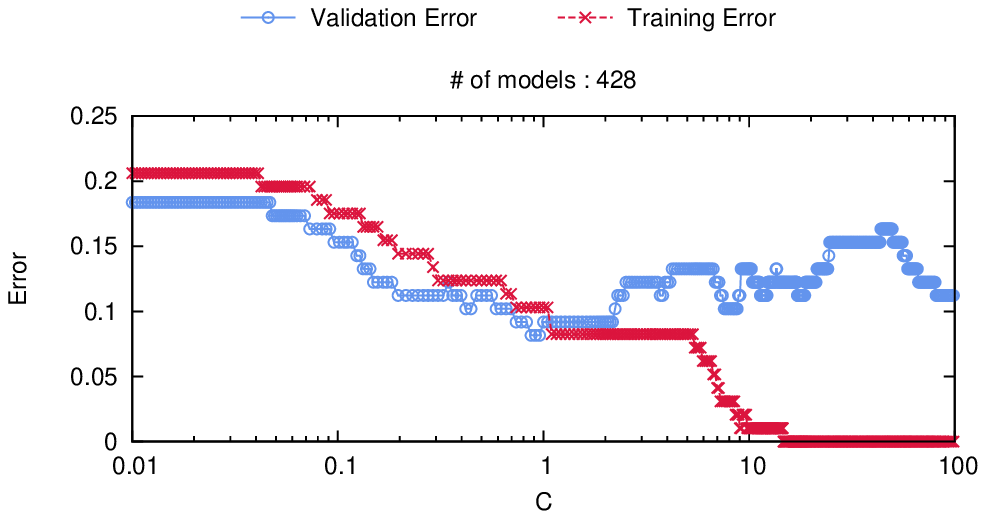} &
   \includegraphics[width=0.45\textwidth]{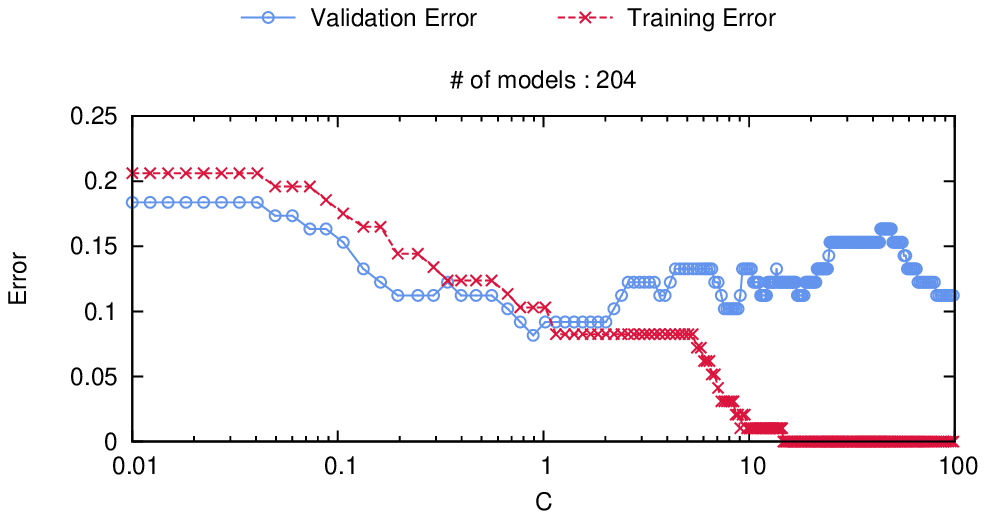} \\
   {\scriptsize (d7) nonlinear SVM with $\veps = 0.05$ on BCP} &
   {\scriptsize (d8) nonlinear SVM with $\veps = 0.1$ on BCP} \\
  \end{tabular}
  \caption{Continued.}
 \end{center}
\end{figure*}


\setcounter{figure}{3}

\begin{figure*}[htbp]
 \begin{center}
  \begin{tabular}{cc}
   \includegraphics[width = 0.45\textwidth]{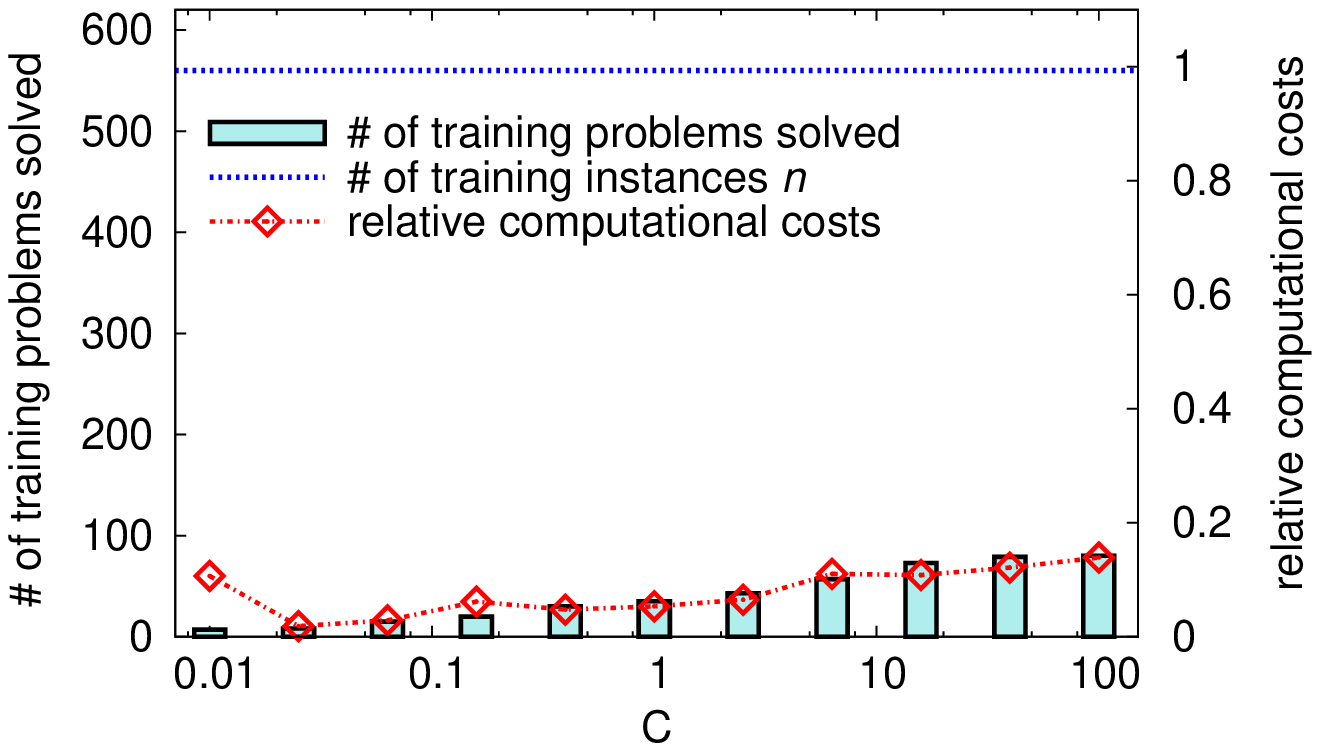} &
   \includegraphics[width = 0.45\textwidth]{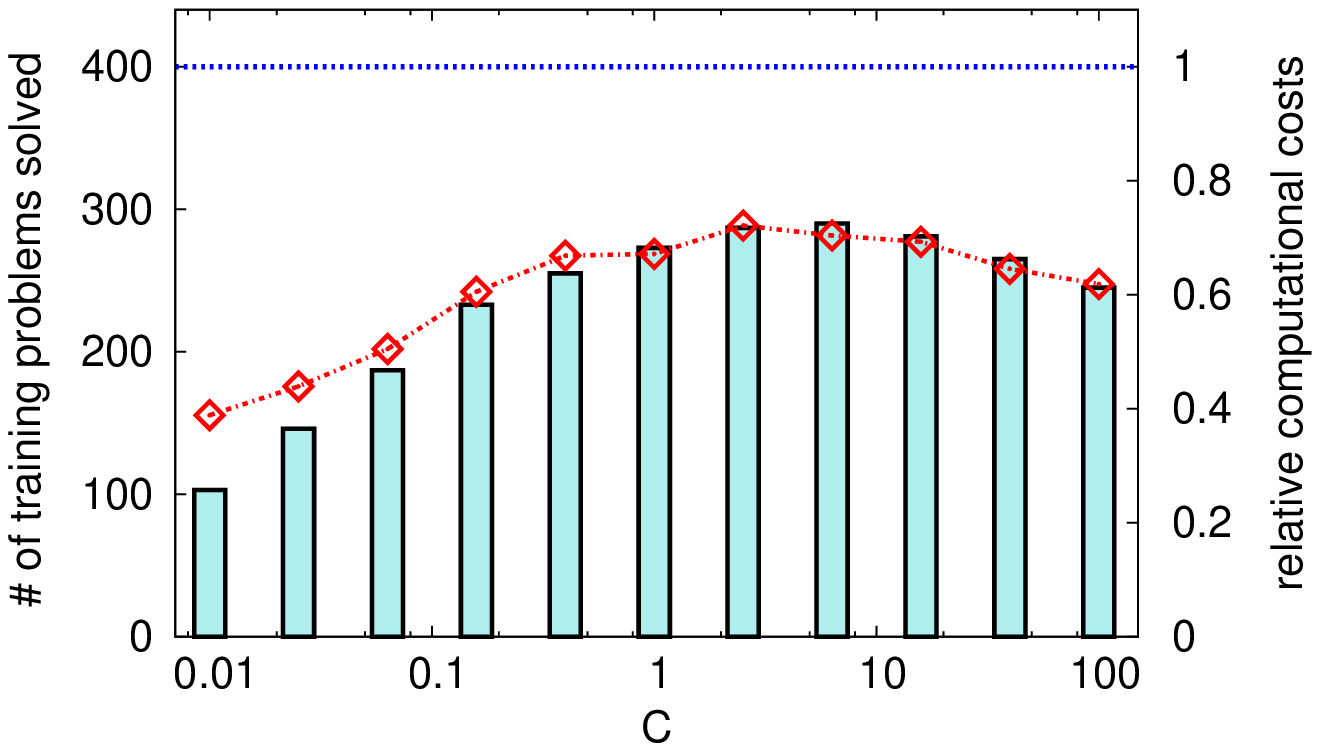} \\
  {\scriptsize (a) BCD} &
  {\scriptsize (b) BCI} \\
   \includegraphics[width = 0.45\textwidth]{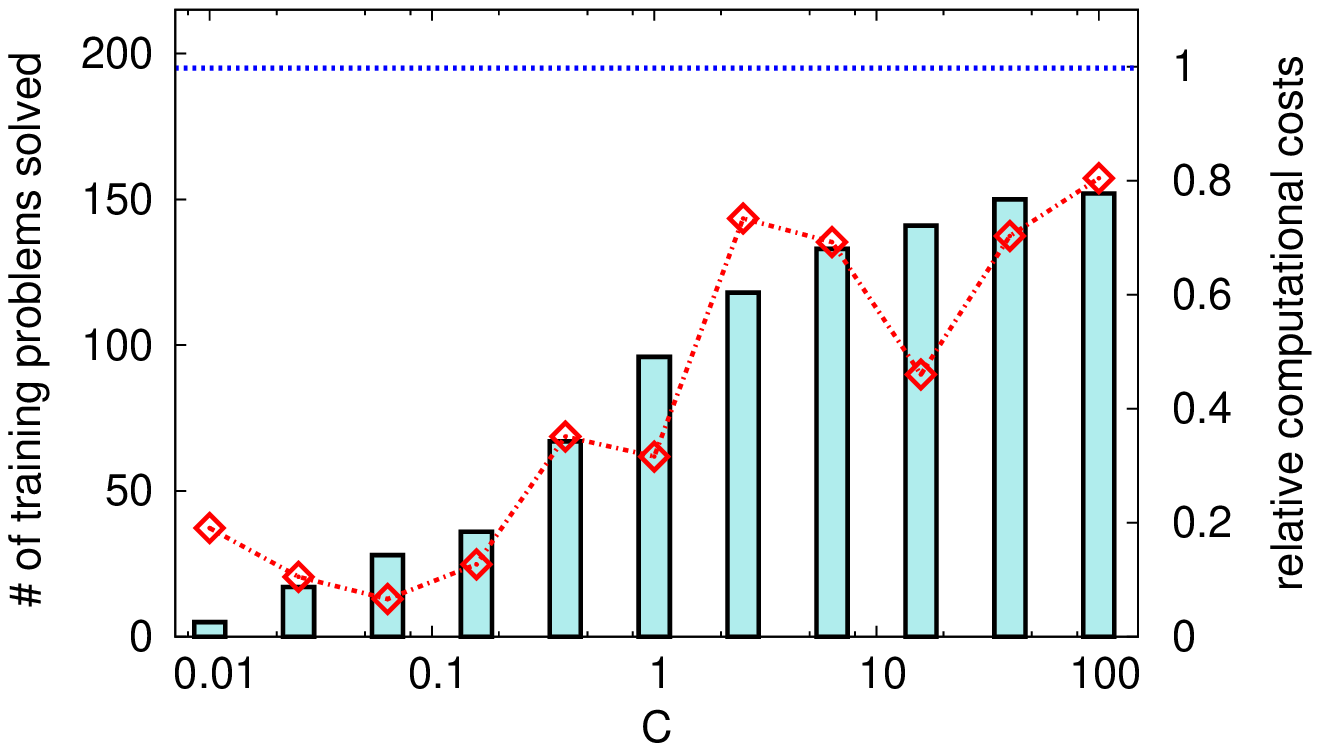} &
   \includegraphics[width = 0.45\textwidth]{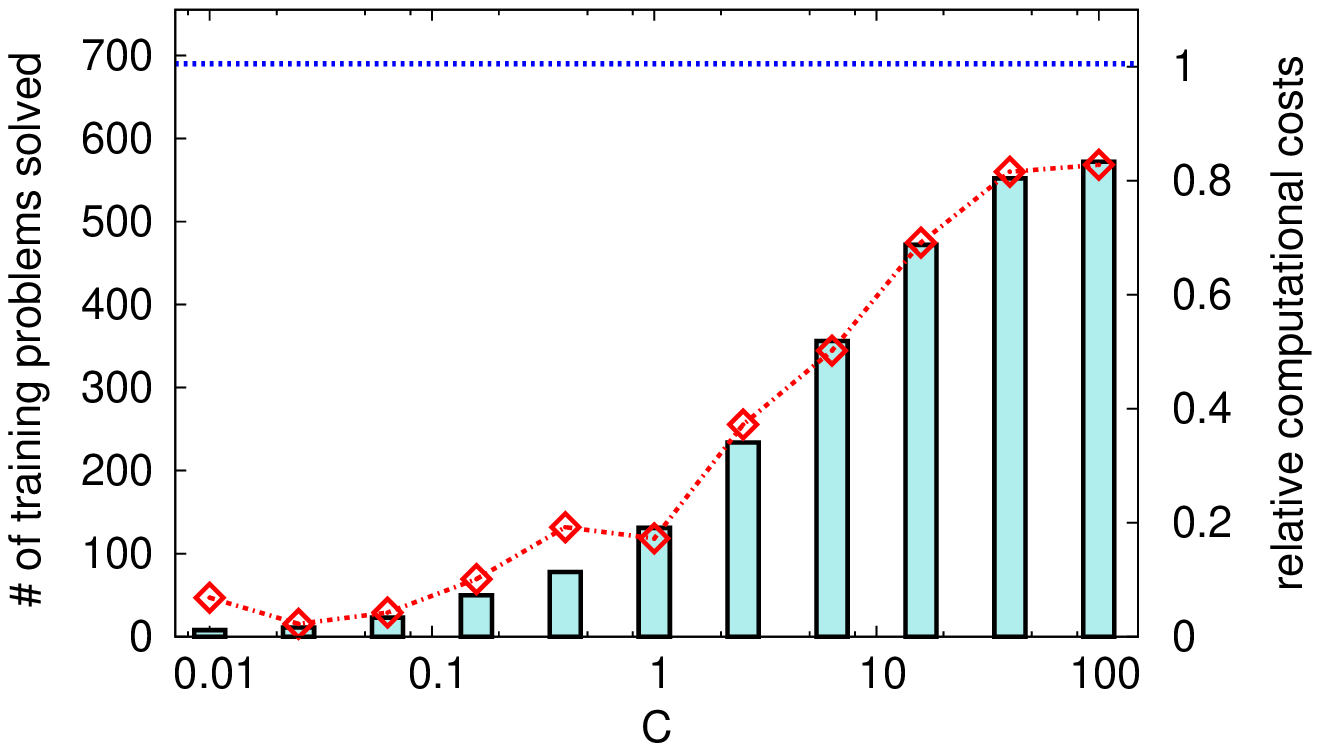} \\
  {\scriptsize (c) PKS} &
  {\scriptsize (d) AUS} \\
   \includegraphics[width = 0.45\textwidth]{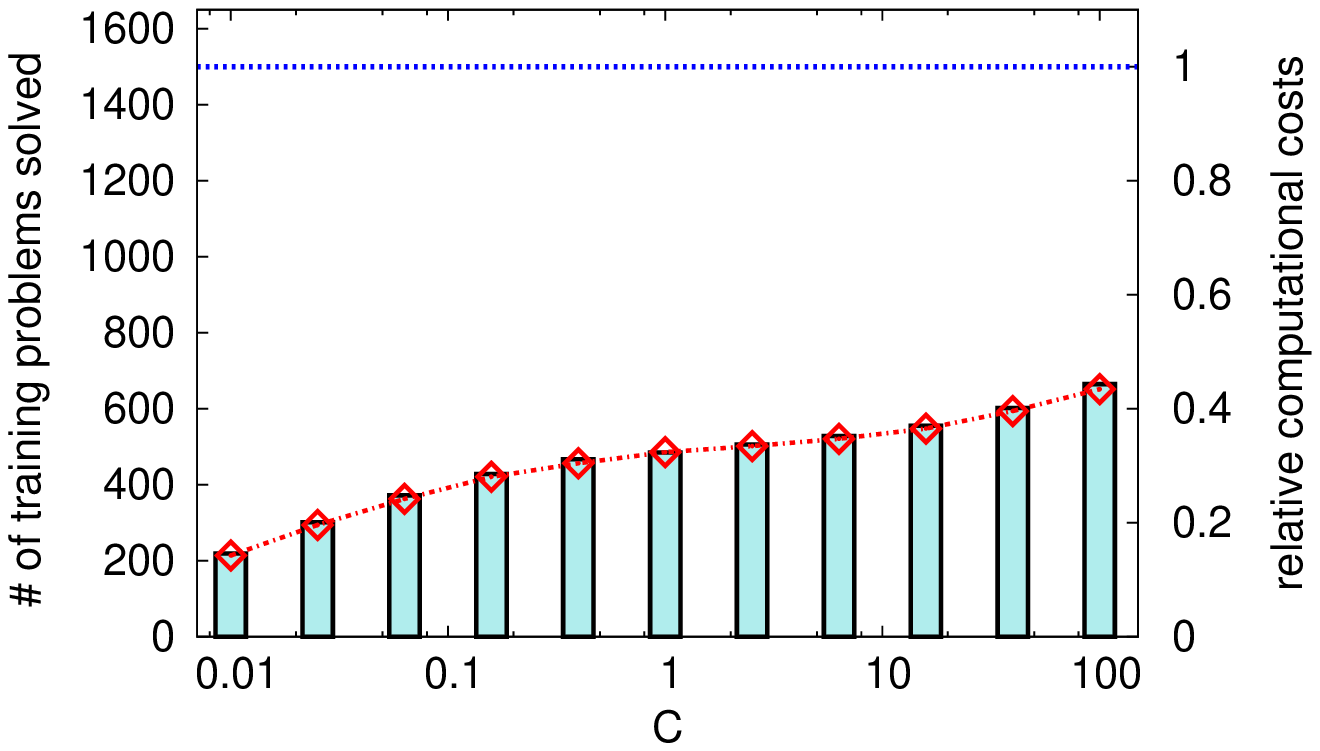} &
   \includegraphics[width = 0.45\textwidth]{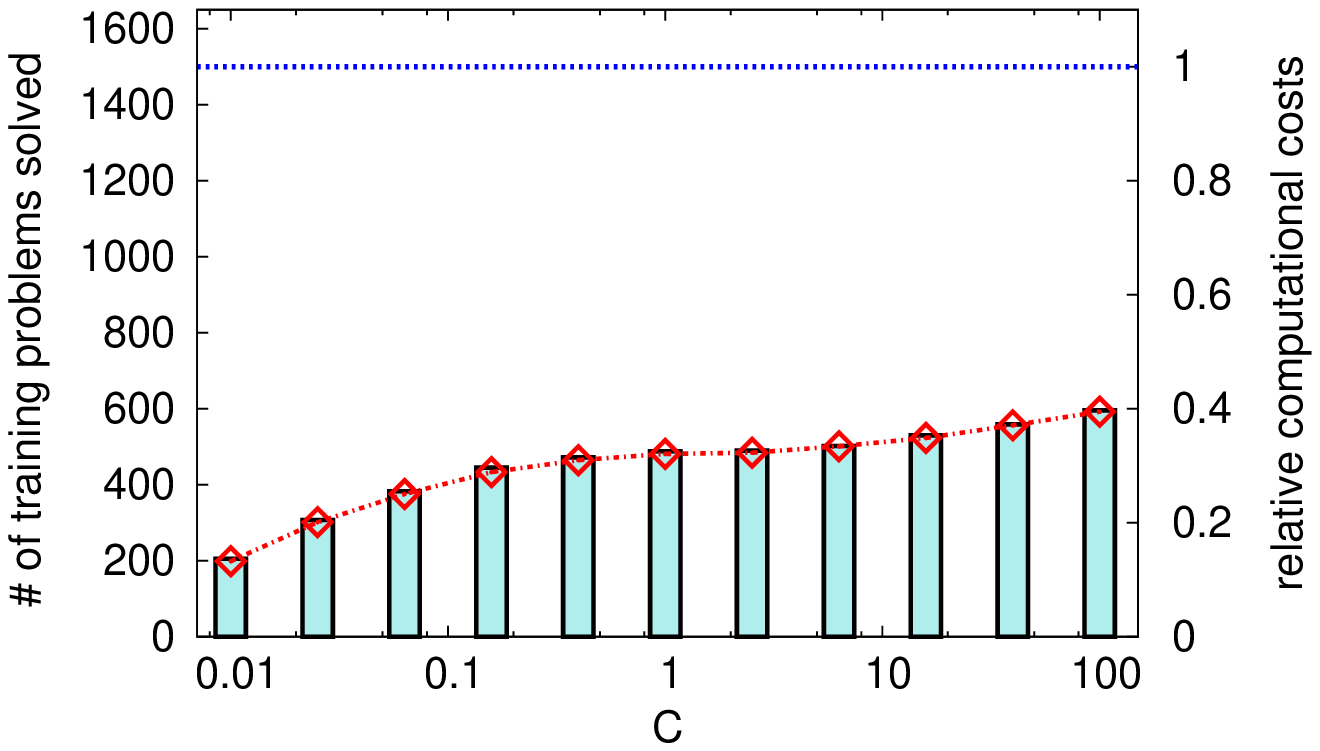} \\
  {\scriptsize (e) G2C} &
  {\scriptsize (f) G2N} \\
   \includegraphics[width = 0.45\textwidth]{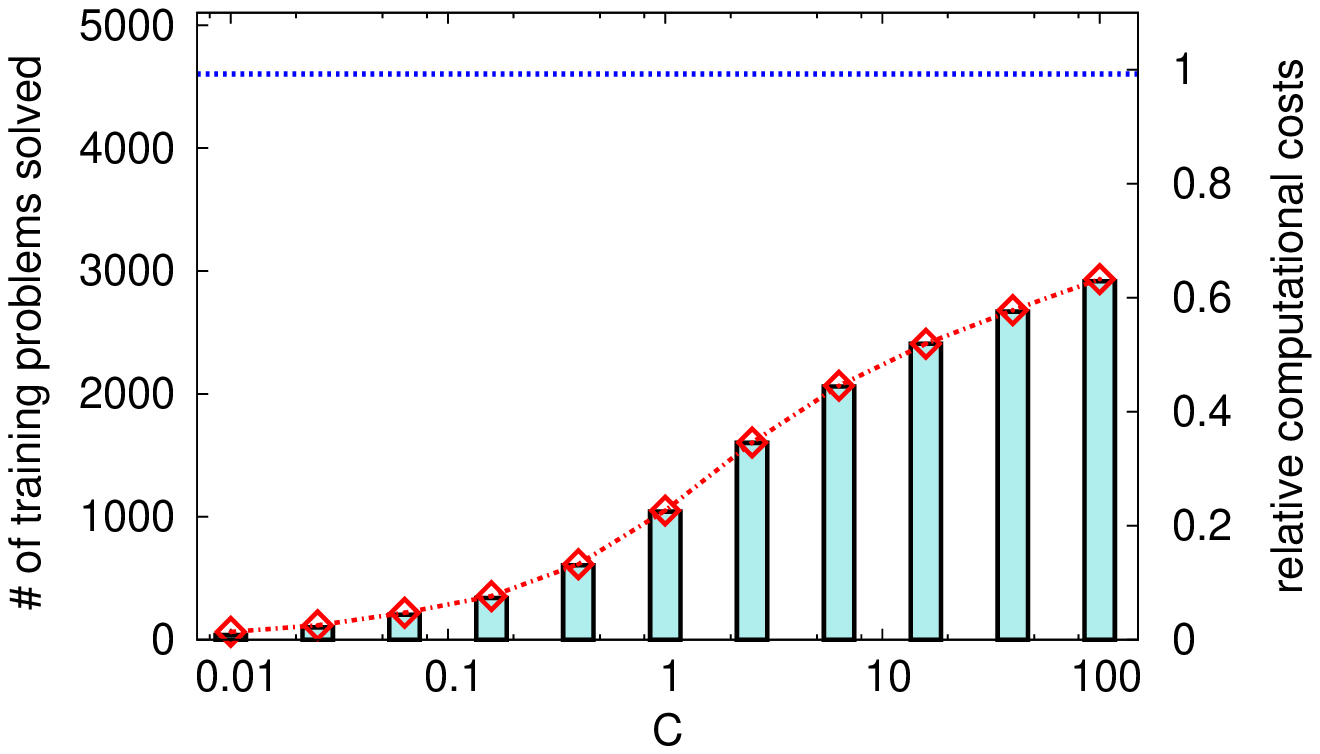} &
   \includegraphics[width = 0.45\textwidth]{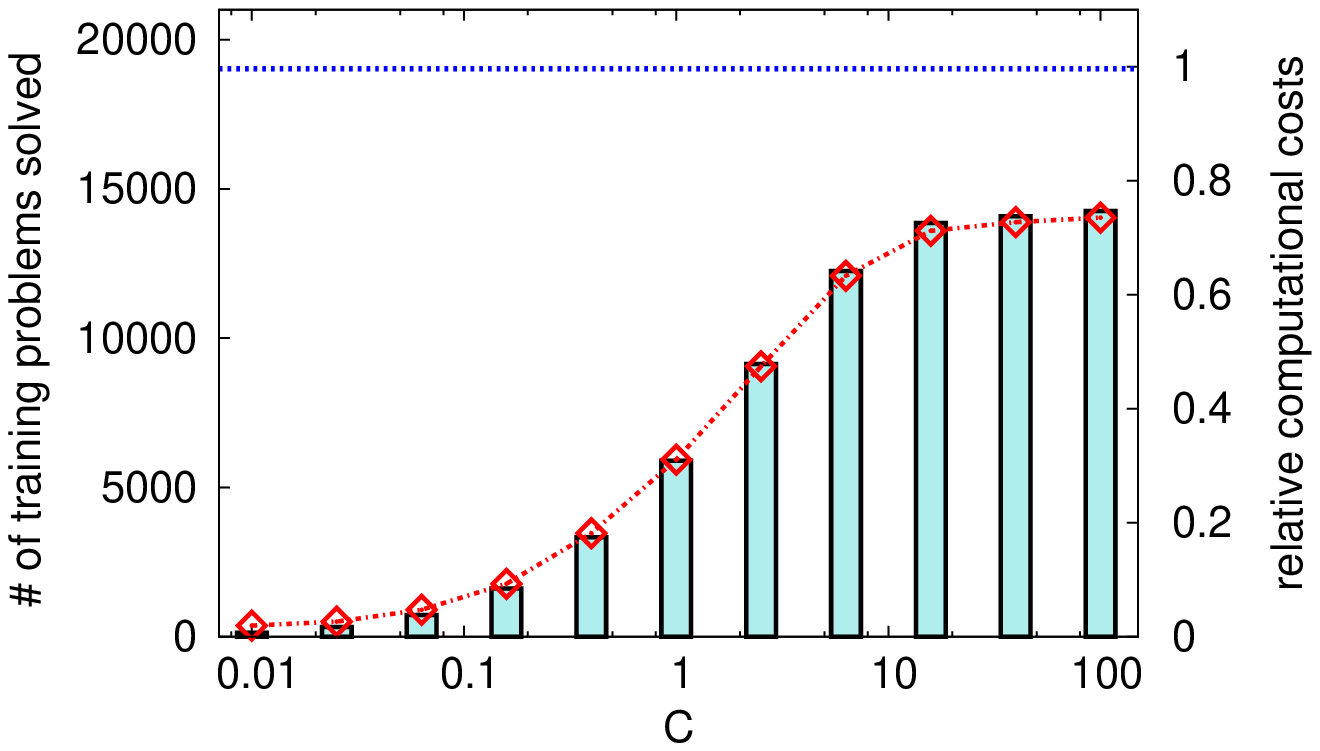} \\
  {\scriptsize (g) SPM} &
  {\scriptsize (h) MGT} 
  \end{tabular}
 \caption{
  The results on fast LOOCV computation experiments.
  The number of optimization problems solved in our proposed approach
  (light blue bars) and the relative computational costs (red dotted lines)
  are shown. 
  }
  \label{fig:LOOCV.result}
 \end{center}
\end{figure*}

\setcounter{figure}{4}

\begin{figure}[htbp]
 \begin{center}
  \begin{tabular}{cc}
   \includegraphics[width = 0.45\textwidth]{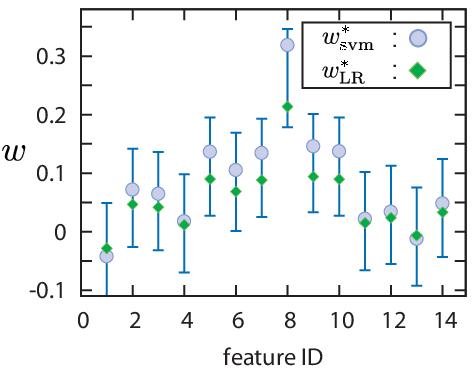} &
   \includegraphics[width = 0.45\textwidth]{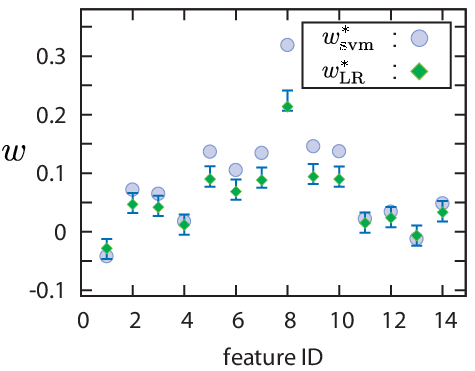} \\
   (a) A single ball &
   (b) Intersection of two balls 
  \end{tabular}
\caption{The lower and the upper bounds of coefficients $w_{\rm LR}^{*}$
  obtained by using the SVM solution as the suboptimal model.
  (a) The bounds were computed based on a single ball in
  Theorem~\ref{theo:main}.
  (b) The bounds were computed based on the intersection of the two balls as
  described in \S~\ref{sec:3}.
  }
\label{fig:LRbySVM}
 \end{center}
\end{figure}


\clearpage
\bibliography{myref}
\bibliographystyle{plain}

\clearpage
\appendix
 \section{Proofs}
\label{app:proofs}


Let us first clarify the optimality condition of a convex constrained optimization problem in the following proposition.
\begin{prop}
 \label{prop:convex.constrained.optimality}
 Consider the following general problem:
 \begin{align}
  \label{eq:general.convex.constrained.problem}
  \min_z ~ g(z) ~~~ {\rm s.t.} ~ z \in \cZ, 
 \end{align}
 where $g$ is a convex differentiable function and $\cZ$ is a convex set. 
 Then a solution $z^*$ is the optimal solution of
 \eq{eq:general.convex.constrained.problem}
 if and only if
 \begin{align*}
  \nabla g(z^*)^\top (z^* - z) \le 0 ~~~ \forall ~ z \in \cZ,
 \end{align*}
 where $\nabla g(z^*)$ is the gradient vector of $g$ at $z = z^*$.
\end{prop}
See, for example, Proposition 2.1.2 in \cite{Bertsekas99a}
for the proof of Proposition~\ref{prop:convex.constrained.optimality}.

\begin{proof}[Proof of Theorem~\ref{theo:main}]
 Let us first rewrite the problem
 \eq{eq:original.problem}
 by using a slack variable
 $\xi \in \RR$
 as
 \begin{subequations}
 \label{eq:alternative.problem}
 \begin{align}
  \label{eq:alternative.problem.a}
  \min_{w \in \cF, \xi \in \RR}
  ~
  &
  J_C (w, \xi) := \frac{1}{2} \|w\|^2
  +
  C \xi
  \\
   \label{eq:alternative.problem.b}
  {\rm s.t.}
  ~~~~
  &
  \xi \ge \sum_{i \in [n]} \ell_i(w).
 \end{align}
\end{subequations}
 It is easy to see that the optimal solution of the problem \eq{eq:alternative.problem} is $w = w^*_C$ and $\xi = \xi^*_C := \sum_{i \in [n]} \ell_i(w^*_C)$.

 Noting that 
 $(\tilde{w}, \tilde{\xi})$
 with
 $\tilde{\xi} := \sum_{i \in [n]} \ell_i(\tilde{w})$
 is a feasible solution of
 \eq{eq:alternative.problem},
 from Proposition~\ref{prop:convex.constrained.optimality},
 \begin{align}
  \label{eq:cond1}
  \nabla J_C(w^*_C, \xi^*_C)^\top
  \left(
  \mtx{c}{
  w^*_C \\
  \xi^*_C
  }
  - 
  \mtx{c}{
  \tilde{w} \\
  \tilde{\xi}
  }
  \right)
  \le 0,
 \end{align}
 where
 $\nabla J_C(w^*_C, \xi^*_C)$
 is the gradient vector of $J_C$ at $(w, \xi) = (w^*_C, \xi^*_C)$.
 By substituting
 $\nabla J_C(w^*_C, \xi^*_C) = (w^{*\top}_C, C)^\top$
 into
 \eq{eq:cond1},
 it is written as the following quadratic inequality constraint:
 \begin{align}
  \label{eq:cond1.a}
  \|w^*_C\|^2 - \tilde{w}^\top w^*_C + C(\xi^*_C - \sum_{i \in [n]} \ell_i(\tilde{w})) \le 0.
 \end{align}
 
 On the other hand, 
 the constraint
 \eq{eq:alternative.problem.b}
 indicates that the optimal solution
 $(w^*_C, \xi^*_C)$
 satisfies the following linear inequality constraint:
 \begin{align}
  \label{eq:cond2}
  \xi^*_C
  \ge
  \sum_{i \in [n]} \ell_i(w^*_C)
  \ge
  \sum_{i \in [n]} \left(\ell_i(\tilde{w}) + \nabla \ell_i(\tilde{w})^\top(w^*_C - \tilde{w}) \right),
 \end{align}
 where
 $\nabla \ell_i(\tilde{w})$
 is the gradient vector 
 (or a subgradient vector in non-differentiable case)
 of $\ell_i$ at $w$.
 Here, note that, 
 the second inequality follows from the assumption that
 $\ell_i$
 is convex,
 and
 the last line is the tangent hyperplane 
 (or a supporting hyperplane in non-differentiable case) 
 of $\ell_i$ at $\tilde{w}$.
 By combining
 \eq{eq:cond1.a}
 and
 \eq{eq:cond2},
 we have
 \begin{align}
  \label{eq:within.ball}
  \| w^*_C - \frac{1}{2} \Bigl(\tilde{w} - C \sum_{i \in [n]} \nabla \ell_i(\tilde{w}) \Bigr) \|^2
  \le
  \Bigl\{ \frac{1}{2} \Bigl\|\tilde{w} + C \sum_{i \in [n]} \nabla \ell_i(\tilde{w})\Bigr\| \Bigr\}^2
  ~\Leftrightarrow~
  \| w^*_C - m \|^2 \le r^2,
 \end{align}
 where
 $m \in \cF$ and $r \ge 0$ are defined in \eq{eq:m.and.r}.

 Since \eq{eq:within.ball} indicates that the optimal solution
 $w^*_C$
 is within the ball
 \begin{align}
  \cS := \{w ~|~ \| w - m \| \le r\},
 \end{align}
 the problem of computing the lower bound of 
 $\theta^\top w^*_C$
 is formulated as
 \begin{align}
  b_{lo}(\theta^\top w^*_C) = \min_{w \in \cS} ~ \theta^\top w.
 \end{align}
 Using the standard Lagrange multiplier theory, 
 the solution of the problem 
 \begin{align*}
  \min_{w}~\theta^\top w~~~{\rm s.t.}~\| w - m\|^2 \le r^2
 \end{align*}
 can be explicitly solved as
 \begin{align*}
  b_{lo}(\theta^\top w^*_C) = \min_{w \in \cS} ~ \theta^\top w = \theta^\top m - \|\theta\| r.
 \end{align*}
 The upper bound of
 $\theta^\top w^*_C$
 is similarly obtained as
 \begin{align*}
  b_{up}(\theta^\top w^*_C) = \max_{w \in \cS} ~ \theta^\top w = \theta^\top m + \|\theta\| r.
 \end{align*}
\end{proof}

%
\begin{proof}[Proof of Theorem 2]
We first consider a case where the loss functions 
$\ell_i, i \in [n]$,
are differentiable
 at $w = w^*_{\tilde{C}}$.
In this case,
we can easily prove the theorem just by substituting 
$w^*_{\tilde{C}}$
into
$\tilde{w}$
and use the proof of Theorem~\ref{theo:main}.
Specifically,
since 
$w^*_{\tilde{C}}$
minimizes
$\frac{1}{2} \|w\|^2 + \tilde{C} \sum_{i \in [n]} \ell_i(w)$,
the gradient at $w = w^*_{\tilde{C}}$ is zero,
i.e.,
 \begin{align}
  \pd{}{w}
  \Bigl(\frac{1}{2} \|w\|^2 + \tilde{C} \sum_{i \in [n]} \ell_i(w) \Bigr)
  \Big|_{w =  w^*_{\tilde{C}}}
  = 0
  ~\Leftrightarrow~
  w^*_{\tilde{C}} + \tilde{C} \sum_{i \in [n]} \nabla \ell_i(w^*_{\tilde{C}}) = 0
  ~\Leftrightarrow~
  \sum_{i \in [n]} \nabla \ell_i(w^*_{\tilde{C}})
  = - \frac{1}{\tilde{C}} w^*_{\tilde{C}}.
 \end{align}
 Thus, 
 in this case,
 the center $m \in \cF$
 and
 the radius $r > 0$
 in
 \eq{eq:m.and.r} 
are written as
 \begin{align}
  \label{eq:new.m.and.r}
  m = \frac{C + \tilde{C}}{2 \tilde{C}} w^*_{\tilde{C}}
  ~\text{and}~
  r = \frac{|C - \tilde{C}|}{2 \tilde{C}} \| w^*_{\tilde{C}}\|.
 \end{align}
 By substituting 
 \eq{eq:new.m.and.r}
 into
 \eq{eq:main.theo.bounds},
 we have the bounds in the form of
 \eq{eq:bounds.model.selection}.

 Next,
 we consider a case where the loss function is not differentiable at
 $w = w^*_{\tilde{C}}$.
 Noting that 
 $w^*_{\tilde{C}}$
 is the optimal solution,
 from Proposition~\ref{prop:convex.constrained.optimality}, 
\begin{align}
\label{eq:VI.on.boundary}
\nabla J_{\tilde{C}}(w_{\tilde{C}}^*, \xi_{\tilde{C}}^*)^\top \bigg(
\bigg[ \begin{array}{c} w_{\tilde{C}}^* \\ \xi_{\tilde{C}}^* \end{array} \bigg] -
\bigg[ \begin{array}{c} \hat{w} \\ \hat{\xi} \end{array} \bigg]
\bigg) \leq 0
~
\text{ for any }
\hat{w} \in \cF,
\end{align}
 where we defined
 $\xi^*_{\tilde{C}} := \sum_{i \in [n]} \ell_i(w^*_{\tilde{C}})$
 and
 $\hat{\xi} := \sum_{i \in [n]} \ell_i(\hat{w})$.
 Since it can be rewritten as
 \begin{align}
  \label{eq:cond2b}
  \sum_{i \in [n]}
  \ell_i(\hat{w})
  \ge
  \sum_{i \in [n]}
  \ell_i(w^*_{\tilde{C}})
  +
  \left(
  - \frac{1}{\tilde{C}} w^*_{\tilde{C}}
  \right)^\top
  (\hat{w} - w^*_{\tilde{C}})
  ~
  \text{ for any }
  \hat{w} \in \cF,
 \end{align}
 we see that 
 $- \frac{1}{\tilde{C}} w^*_{\tilde{C}} \in \cF$
 is a subgradient vector of
 $\sum_{i \in [n]} \ell_i(w)$
 at
 $w = w^*_{\tilde{C}}$,
 i.e.,
 we can replace
 $\sum_{i \in [n]} \nabla \ell_i(\tilde{w})$
 in
 \eq{eq:cond2}
 with
 $- \frac{1}{\tilde{C}} w^*_{\tilde{C}}$
 when
 $\tilde{w} = w^*_{\tilde{C}}$.
 If we set 
 $\hat{w} := w^*_{C}$
 in \eq{eq:cond2b},
 we have the following linear inequality constraint on $w^*_C$:
 \begin{align}
  \label{eq:cond2a}
  \xi^*_C
  \ge
  -\frac{1}{\tilde{C}} w^{*\top}_{\tilde{C}} w^*_C
  +
  \frac{1}{\tilde{C}} \| w^*_{\tilde{C}} \|^2
  +
  \sum_{i \in [n]} \ell_i(w^*_{\tilde{C}}).
 \end{align}
 By combining
 \eq{eq:cond1.a}
 with $\tilde{w} := w^*_{\tilde{C}}$
 and
 \eq{eq:cond2a},
 we have
 \begin{align}
  \label{eq:within.ball.2}
  \| w^*_C - \frac{C + \tilde{C}}{2 \tilde{C}} w^*_{\tilde{C}} \|^2
  \le
  \left\{ \frac{|C - \tilde{C}|}{2 \tilde{C}} \| w^*_{\tilde{C}}\| \right\}^2.
 \end{align}
 It indicates that the optimal solution
 $w^*_C$
 is in the ball with the center and the radius defined by
 \eq{eq:new.m.and.r}.
\end{proof}

\section{Additional Theoretical Results}
\label{app:additional.theory}
%
%
\paragraph{Bounding Lasso dual solutions}
We can easily confirm that the lower and the upper bounds
\eq{eq:main.theo.bounds} 
in Theorem~\ref{theo:main}
are still true 
when we have some additional convex constraints in \eq{eq:original.problem}.
The following theorem tells that
we can obtain similar bounds for LASSO problem.
\begin{theo}
\label{theo:App.to.LASSO}
 Let us consider a regression problem with the training set
 $\{(x_i, y_i)\}_{i \in [n]}, x_i \in \RR^d, y_i \in \RR$.
 We denote the $n \times d$ input (design) matrix as 
 $X := [z_1~\ldots~z_d] \in \bR^{n \times d}$,
 where
 $z_j$
 represents the $j^{\rm th}$ column of $X$,
 and the $n$-dimensional output (target) vector as
 $y := [y_1, \ldots, y_n]^\top \in \RR^n$.

 A well-known Lasso problem is formulated as 
 \begin{align}
  \label{eq:Primal.LASSO}
  \beta^*_{\lambda} := \arg \min_{\beta \in \RR^d}
  ~
  \frac{1}{2} \| y - X \beta\|^2 + \lambda \| \beta\|,
 \end{align}
 where
 $\lambda > 0$
 is the regularization parameter.
 The dual of \eq{eq:Primal.LASSO}
 is written as 
  \begin{align}
   \label{eq:Dual.LASSO}
   \alpha_\lambda^*
   &
   := \arg \min_{w \in \bR^n}
   ~
   \frac{1}{2} \| \alpha - \frac{1}{\lambda} y \|^2
    ~~~
   {\rm s.t. }
   ~
   \| X^\top \alpha \|_\infty \le 1,
  \end{align}
 where
 $\alpha \in \RR^n$
 is the Lagrange multipliers.
 Then, 
 for any
 $\theta \in \RR^n$,
 the inner product
 $\theta^\top \alpha^*_\lambda$
 is bounded as 
 \begin{eqnarray*}
  \theta^\top m_{\rm Lasso} - \|\theta\| r_{\rm Lasso} 
   \le 
   \theta^\top \alpha_\lambda^*
   \le 
   \theta^\top m_{\rm Lasso} + \|\theta\| r_{\rm Lasso},
 \end{eqnarray*} 
 where
 $m_{\rm Lasso} \in \RR^d$
 and
 $r_{\rm Lasso} > 0$
 are defined,
 with any $\tilde{\alpha} \in \RR^n$, 
 as
 \begin{align}
  m_{\rm Lasso} := \frac{1}{2} (\tilde{\alpha} + \frac{1}{\lambda} y),
  ~~
  r_{\rm Lasso} := \frac{1}{2} \| \tilde{\alpha} - \frac{1}{\lambda} y \|.
 \end{align}
\end{theo}
 \begin{proof}
  The dual problem
  \eq{eq:Dual.LASSO}
  is rewritten as 
  \begin{align}
   \alpha_\lambda^*
   :=
   \arg \min_{\alpha \in \bR^n}
   ~
   \frac{1}{2} \| \alpha - \frac{1}{\lambda} y \|^2
   ~~~
   {\rm s.t. }
   ~
   \| X^\top \alpha \|_\infty \le 1
   \label{eq:Lasso.dual.2}
   =
   \arg \min_{\alpha \in \bR^n}
   ~
   \frac{1}{2} \| \alpha \|^2
   -
   \frac{1}{\lambda}
   \sum_{i \in [n]}
   \alpha_i y_i
   ~~~
   {\rm s.t. }
   ~
   \| X^\top \alpha \|_\infty \le 1.
  \end{align}
  Noting that
  \eq{eq:Lasso.dual.2}
  has the same form as our
  $L_2$ regularized learning problem
  in
  \eq{eq:original.problem},
  we can similarly compute the lower and the upper bounds of the inner product
  $\theta^\top \alpha^*_{\lambda}$.
\end{proof}

An important consequence of
Theorem~\ref{theo:App.to.LASSO}
is that,
the lower and the upper bounds of the (negative) residual of the Lasso 
$x_i^\top \beta^*_\lambda - y_i, i \in [n]$, 
can be obtained by using the relationship: 
\begin{align}
 (\alpha^*_\lambda)_i \equiv f(x_i) - y_i, ~ i \in [n].
\end{align}
Specifically,
the residual is bounded as
\begin{align}
 - (m_{\rm Lasso})_i - r_{\rm Lasso}
 \le
 y_i - x_i^\top \beta^*_\lambda
 \le 
 - (m_{\rm Lasso})_i + r_{\rm Lasso}.
\end{align}

Another important relationship in Lasso is
\begin{align}
 | z_j^\top \alpha^*_\lambda | < 1
 ~\Rightarrow~
 (\beta^*_{\lambda})_j = 0,
 ~
 j \in [d].
\end{align}
Using our bounds,
it indicates that
\begin{align}
 \max \biggl\{
 \bigg| z_j^\top m_{\rm Lasso} - \|z_j\| r_{\rm Lasso} \bigg|, 
 \bigg| z_j^\top m_{\rm Lasso} + \|z_j\| r_{\rm Lasso} \bigg|
 \biggr\}
 <
 1
 ~\Rightarrow~
(\beta^*_{\lambda})_j = 0,
\end{align}
i.e.,
the
$j^{\rm th}$
variable can be removed without actually computing the optimal solution $\beta^*_\lambda$. 
This computational trick is called
\emph{safe screening}
and has been intensively studied in the
literature~\cite{ElGhaoui12b,Xiang12a,Xiang12b,Dai12a,Wang12a,Ogawa13a,Wang13c,Wang13d,Wu13a,Wang13a,Wang13b,Ogawa14a}.
Actually,
we can easily show that our ball defined in 
Theorem~\ref{theo:App.to.LASSO} is equivalent to
(14) in \cite{Liu13a}. 
In this sense, 
our results in
Theorems~\ref{theo:main}
and
\ref{theo:App.to.LASSO}
are considered as the general form that includes safe screening as a special case.

\paragraph{How to find small intersection of two balls}
In \S~\ref{sec:3},
we slightly mentioned about a simple trick for finding a small intersection of two balls
in which the optimal solution
$w^*_C$ 
is guaranteed to exist.
When we have a suboptimal model 
$\tilde{w} \in \cF$,
our idea is to make use of the center $m \in \cF$ in
\eq{eq:m}
as another suboptimal solution,
and consider the intersection of the resulting two balls.
The following lemma indicates that the volume of the intersection is at most half of the original ball. 
\begin{lemm}
\label{lemm:Recursive.IT}
 For any $\tilde{w} \in \cF$,
 let
 $\{ \tilde{w}_t \in \cF\}_{t \in \bN}$
 be the series of vectors defined by
 \begin{eqnarray*}
  \tilde{w}_1 := \tilde{w}
  ~\text{and}~
  \tilde{w}_{t+1} = \frac{1}{2} \left(\tilde{w}_t - C \sum_{i\in[n]} \nabla \ell_i(\tilde{w}_t) \right)~
  \forall t \ge 1.
 \end{eqnarray*}
 Furthermore,
 let
 $\cS(w)$
 be the ball obtained by 
 Theorem~\ref{theo:main} 
 when we used 
 $\tilde{w}$ 
 as the suboptimal solution.
 Then, 
 $\{ \tilde{w}_t \}_{t \in \bN}$ 
 satisfy the following property:
\begin{eqnarray}
{\rm Vol}\left(\cS(\tilde{w}_{t+1}) \cap \cS(\tilde{w}_t) \right)
< 
\frac{1}{2} {\rm Vol}(\cS(\tilde{w}_t))
~\forall t \in \bN,
\end{eqnarray}
where 
${\rm Vol}(\cS)$ 
indicates the volume of 
$\cS$. 
\end{lemm}
\begin{proof}[Proof of Lemma~\ref{lemm:Recursive.IT}]
By Theorem~\ref{theo:main}, 
the center 
$m_t$ 
and the radius 
$r_t$ 
of the ball 
$\cS(\tilde{w}_t)$ 
are written as 
\begin{eqnarray*}
m_t &=& \frac{1}{2} \left( \tilde{w}_t - C \sum_{i\in [n]} \nabla\ell_i(\tilde{w}_t) \right) = \tilde{w}_{t+1},
\\
r_t &=& \frac{1}{2} \left\| \tilde{w}_t + C \sum_{i\in [n]} \nabla\ell_i(\tilde{w}_t) \right\|. 
\end{eqnarray*}
Then, 
$\forall t \in \NN$,
\begin{eqnarray*}
\|m_{t+1} - m_t \|^2 
=
\| \tilde{w}_{t+2} - \tilde{w}_{t+1} \|^2 
=
\left\| - \frac{1}{2} \left( \tilde{w}_{t+1} + C \sum_{i\in [n]}\nabla \ell_i(\tilde{w}_{t+1}) \right) \right\|^2
=
r_{t+1}^2. 
\end{eqnarray*}
It indicates that 
the center 
$m_t$ 
is on the hypersphere of 
$\cS(\tilde{\bm w}_{t+1})$, 
i.e., 
there exists a half space 
$\cH_t$ 
whose boundary is the tangent hyperplane of 
$\cS(\tilde{\bm w}_{t+1})$
 at $m_t$.
Using 
$\cH_t$,
 we can show that 
\begin{eqnarray*}
{\rm Vol}\left(\cS(\tilde{\bm w}_{t+1}) \cap \cS(\tilde{\bm w}_t) \right)
<
{\rm Vol}\left(\cH_t \cap \cS(\tilde{\bm w}_t) \right)
=
\frac{1}{2} {\rm Vol}(\cS(\tilde{\bm w}_t)).
\end{eqnarray*}
\end{proof}
Note that 
Lemma~\ref{lemm:Recursive.IT}
holds for any loss functions
$\{\ell_i\}_{i \in [n]}$. 
Thus,
once we construct a ball including 
$w_C^*$ 
as in Theorem~\ref{theo:main}, 
we can reduce the volume of the closed convex domain $\cS$ without any additional information, 
and it enables us to obtain tighter bounds. 

\end{document}